\documentclass[10pt,twocolumn,letterpaper]{article}

\usepackage{cvpr}              % To produce the CAMERA-READY version
\definecolor{cvprblue}{rgb}{0.21,0.49,0.74}
\usepackage[pagebackref,breaklinks,colorlinks,citecolor=cvprblue]{hyperref}
\usepackage{colortbl} % For table colors
\usepackage{multirow} % For multi-row cells
\usepackage{booktabs} % For top/mid/bottom rule lines

\usepackage[capitalize]{cleveref}
\crefname{section}{Sec.}{Secs.}
\Crefname{section}{Section}{Sections}
\Crefname{table}{Table}{Tables}

\def\paperID{1549} % *** Enter the Paper ID here
\def\confName{CVPR}
\def\confYear{2025}

\title{LUCAS: Layered Universal Codec Avatars}
\vspace{-30pt}
\author{Di Liu$^{1,2}$ \quad Teng Deng$^1$ \quad Giljoo Nam$^1$ \quad Yu Rong$^1$ \quad Stanislav Pidhorskyi$^1$ \quad Junxuan Li$^1$ \\
\quad Jason Saragih$^1$ \quad Dimitris N. Metaxas$^2$ \quad Chen Cao$^1$ \vspace{+0.3em} \\
$^1$Codec Avatars Lab, Meta~~~$^2$Rutgers University\vspace{-0em} \\
}
\begin{document}
\twocolumn[{%
\renewcommand\twocolumn[1][]{#1}%
\maketitle
\vspace{-30pt}
\begin{center}
    \centering
\includegraphics[width=1\linewidth]{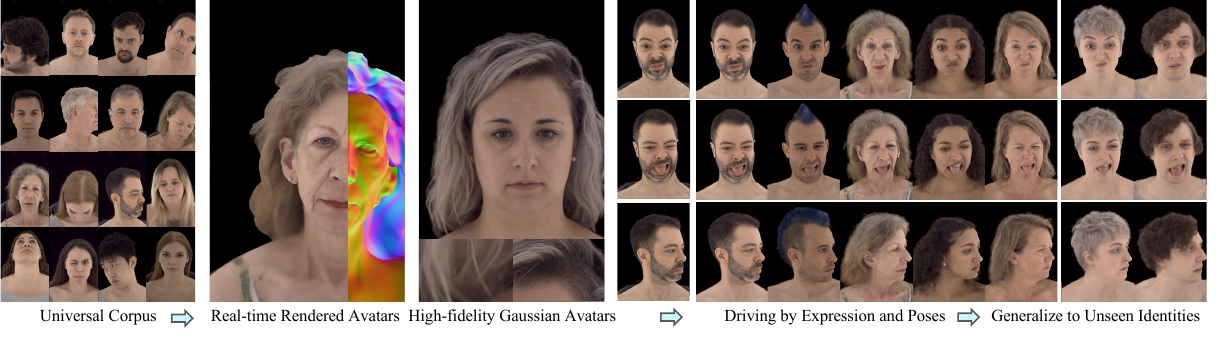}
\vspace{-20pt}
\captionof{figure}{
\textbf{LUCAS}: A novel approach for high-fidelity \textbf{L}ayered \textbf{U}niversal \textbf{C}odec \textbf{A}vatar\textbf{s}. We disentangle face and hair into a layered structure, supporting both real-time mesh-based avatar~(45 FPS on mobile) and high-fidelity Gaussian avatar generation. Our universal layered prior model also enables accurate expression and pose transfer, even for unseen subjects, while maintaining visual quality.
}

\label{cover_image}
\end{center}%
}]

\begin{abstract}
\vspace{-23pt}
% Original version
% We present LUCAS, a novel 3D universal head avatar representation that disentangles the face and hair geometry and appearance into a layered structure. Unlike previous single-mesh avatars, LUCAS's compositional design enables more accurate hair dynamics and alignment during head movements and expressions.
%LUCAS introduces a Universal Layered Prior Model that powers both the face and hair branches, enabling cross-identity generalization while maintaining real-time performance. Our model can further integrate with Gaussian splatting to achieve higher visual fidelity at the cost of computational efficiency, particularly beneficial for complex hairstyles.
%Experiments demonstrate that LUCAS outperforms previous single-mesh and Gaussian-based avatar models in both quantitative and qualitative evaluations, including on held-out subjects in zero-shot driving scenarios. LUCAS exhibits enhanced dynamic performance in handling head pose changes, expression transfer, and hairstyle variations, advancing the state-of-the-art in 3D head avatar reconstruction with flexible quality-performance trade-offs.

% Chen's version
Photorealistic 3D head avatar reconstruction faces critical challenges in modeling dynamic face-hair interactions and achieving cross-identity generalization, particularly during expressions and head movements.
We present LUCAS, a novel Universal Prior Model (UPM) for codec avatar modeling that disentangles face and hair through a layered representation. Unlike previous UPMs that treat hair as an integral part of the head, our approach separates the modeling of the hairless head and hair into distinct branches.
LUCAS is the first to introduce a mesh-based UPM, facilitating real-time rendering on devices.
Our layered representation also improves the anchor geometry for precise and visually appealing Gaussian renderings. Experimental results indicate that LUCAS outperforms existing single-mesh and Gaussian-based avatar models in both quantitative and qualitative assessments, including evaluations on held-out subjects in zero-shot driving scenarios. LUCAS demonstrates superior dynamic performance in managing head pose changes, expression transfer, and hairstyle variations, thereby advancing the state-of-the-art in 3D head avatar reconstruction. \href{https://lsn33096.github.io/LUCAS/}{Project page: https://lsn33096.github.io/LUCAS/.}

\end{abstract}    
\vspace{-10pt}
\section{Introduction}
\label{intro}

Photorealistic 3D head avatars are vital for authentic communication in virtual and augmented environments, where capturing subtle expressions and head movements is crucial~\cite{lombardi2018deep,wei2019vr,richard2021audio}. High-quality avatars enhance experiences in telecommunications, social VR, virtual training, and healthcare by ensuring accurate geometry and appearance, particularly in dynamic scenarios involving facial and hair deformations~\cite{chu2020expressive,schwartz2020eyes,liu2021label,liu2021refined,gao2022data,zhangli2022region,chang2022deeprecon,liu2022transfusion,he2023dealing,martin2023deep,gao2024training,liu2024lepard,liu2023deformer,liu2023deep,liu2024Instantaneous,he2024dice,han2024proxedit,zhangli2024layout,dao2025improved}.
Recent advances in Codec Avatars~\cite{lombardi2021mixture,li2023megane} have achieved remarkable photorealism through sophisticated rendering techniques and volumetric primitives. But these methods often demand significant computational resources, posing challenges for real-time rendering on mobile devices.
%Recent advances in Codec Avatars \cite{lombardi2018deep, lombardi2021mixture}
%have achieved remarkable photorealism through sophisticated rendering techniques and volumetric primitives. However, these methods often require significant computational resources, making real-time rendering on mobile devices challenging.

Pixel Codec Avatars (PiCA)~\cite{ma2021pixel} tackle performance challenges by introducing a pixel-level decoder that dynamically adjusts texture resolution in screen space. This approach enables efficient real-time rendering on mobile devices through direct per-pixel color decoding, eliminating the need for fixed texture maps or vertex-based representations. However, PiCA is a personalized model that requires time-consuming per-identity training, which limits its scalability. Additionally, its single-mesh representation struggles with accurately reconstructing hair, often resulting in artifacts such as hair tails appearing on shoulders or unnatural hair deformation during head movements.

% Universal avatar reconstruction approaches, such as~\cite{cao2022authentic} and URAvatar~\cite{li2024uravatar}, 
% Universal avatar reconstruction approaches~\cite{cao2022authentic,li2024uravatar} have made progress in cross-identity generalization, allowing the codec avatar generalzed from the universal prior model trained on multiple users data. However, these methods face significant challenges in hair modeling and dynamic scenarios. Their simplistic representations and inaccurate guide meshes often result in misaligned geometry and limited hair modeling capabilities, preventing natural deformation during expressions and head movements.

Universal avatar reconstruction approaches~\cite{cao2022authentic,li2024uravatar} have advanced cross-identity generalization, enabling codec avatars to generalize from universal prior models (UPM) trained on data from multiple users. However, these methods encounter significant challenges in hair modeling and dynamic scenarios. Their simplistic representations and inaccurate guide meshes often lead to misaligned geometry and limited hair modeling capabilities, hindering natural deformation during expressions and head movements.

%To address these challenges, we propose \textbf{LUCAS}, \textbf{L}ayered \textbf{U}niversal \textbf{C}odec \textbf{A}vatar\textbf{s}. LUCAS introduces a layered representation that separates face and hair components, allowing them to deform independently while ensuring precise alignment. This design enables more accurate hair dynamics—using the same encoding features but decoding them separately for face and hair. For instance, in Fig.~\ref{fig:vis_frown}, when the subject gazes upward and frowns, the hair naturally lowers toward the eyebrows. While a single-mesh representation struggles with such subtle movements due to uniform expression code influence, LUCAS's layered structure ensures independent and realistic face and hair deformation.
%
% To address these challenges, we propose \textbf{LUCAS}, or \textbf{L}ayered \textbf{U}niversal \textbf{C}odec \textbf{A}vatar\textbf{s}, a layered representation that separates face and hair components for independent deformation while maintaining precise alignment.  This design enables more accurate hair dynamics by using the same encoding features but decoding them separately for the face and hair. For instance, as shown in Fig.~\ref{fig:vis_frown}, when the subject gazes upward and frowns, the hair naturally lowers toward the eyebrows. While a single-mesh representation is not flexible and cannot faithfully disentangle the movement of face and hair as separate factors, LUCAS's layered structure ensures their independent and realistic deformation.

\begin{figure}[t]
% \vspace{-15pt}
\centering
\includegraphics[width=1\linewidth]{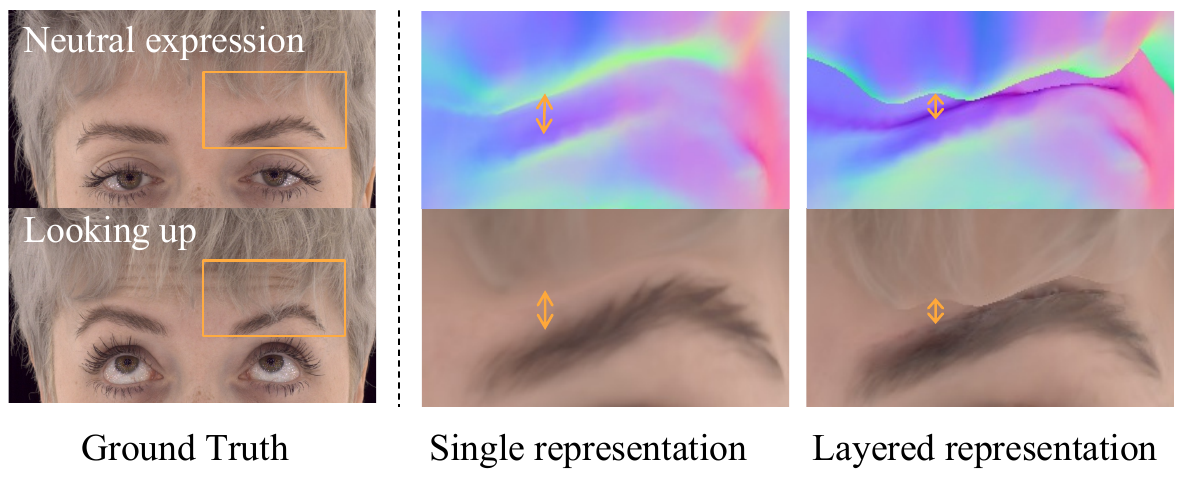}
\vspace{-20pt}
\caption{
\textbf{Layered representation enables adaptive alignment between face and hair.} LUCAS's independent face and hair deformation captures subtle hair movements in response to facial expressions, unlike single-mesh avatars that are globally controlled.
}
\label{fig:vis_frown}
\vspace{-20pt}
\end{figure}

To address these challenges, we propose \textbf{LUCAS} (\textbf{L}ayered \textbf{U}niversal \textbf{C}odec \textbf{A}vatar\textbf{s}), a layered representation that separates face and hair components, allowing them to deform independently while maintaining precise alignment. This design enables more accurate hair dynamics by using shared encoding features but decoding them separately for the face and hair. For instance, as shown in Fig.~\ref{fig:vis_frown}, when the subject gazes upward and frowns, the hair naturally lowers toward the eyebrows. In contrast, single-mesh representations lack the flexibility to disentangle face and hair movements as separate factors, leading to interdependent deformations. Smoothness regularization further exacerbate this issue by enforcing coupled motion. LUCAS overcomes these limitations, enabling realistic and independent deformation for natural movement.
LUCAS follows a universal training strategy, training the UPM on data from multiple users, which allows it to generalize easily to unseen users and generate realistic codec avatars. 
% Additionally, LUCAS can optionally integrate 3D Gaussian Splatting with its layered structure to enhance the fidelity of reconstructed avatars, albeit at the cost of computational efficiency.
Additionally, we show that our layered mesh design improves the anchor geometry for precise and visually appealing Gaussian renderings~\cite{kerbl20233d}.
% Moreover, we propose a Universal Layered Prior Model for our layered approach, enabling cross-identity scalability while maintaining real-time efficiency. Additionally, LUCAS can optionally integrate 3D Gaussian Splatting~\cite{kerbl20233d} with the layered structure to improve the fidelity of reconstructed avatars at the cost of computational efficiency. Our contributions are:
In summary, our contributions are:
\begin{itemize}
% \item We present LUCAS, the first codec avatar that unifies universal modeling and layered representation for face and hair, enabling both cross-identity generalization and improved face-hair alignment while maintaining real-time performance on device.
% \item LUCAS's architecture allows for optional integration with Gaussian splatting to achieve higher visual fidelity, particularly for complex hairstyles.
% \item LUCAS exhibits enhanced dynamic performance in handling head pose changes, expression transfer, and hairstyle variations, including on held-out subjects in zero-shot driving scenarios.

\item We introduce LUCAS, the first mesh-based Universal Prior Model that enables cross-identity generalization while maintaining real-time rendering on devices.
\item The first compositional Universal Prior Model for the head, featuring a layered representation for both the hairless head and hair, which significantly enhances the quality of both face and hair rendering.
\item LUCAS demonstrates improved dynamic performance in managing head pose changes, expression transfers, and hairstyle variations, even on unseen subjects in zero-shot driving scenarios.

\end{itemize}

% \cc{
% I think the structure of introduction should be:
% \begin{itemize}
%     \item Codec Avatar (HQLP, MVP): +photo-realistic, -slow in rendering on device
%     \item PiCA: + fast rendering speed, high-precise geometry; - person-specific
%     \item Universal avatar: + universal, generalize to unseen id; - cannot handle hair well
%     \item Our method: + universal, layered, handle hair etc.
% \end{itemize}
% }

% ~\di{I rewrote the intro section following chen's suggestion. Giljoo please check.}

\section{Related Works}
\noindent \textbf{3D Head Avatar Reconstruction.}
Early approaches to head avatar reconstruction were primarily based on 3D Morphable Face Models (3DMFMs)~\cite{blanz2023morphable}, which used linear combinations of prototype vectors for shape and texture generation, later extended with blendshapes for animation~\cite{lewis2014practice}. However, these manual blendshape-based approaches were limited in expressiveness and required significant effort to create. Deep learning has revolutionized this field, introducing non-linear models through VAEs~\cite{ranjan2018generating} and GANs~\cite{shamai2019synthesizing} for more complex facial representations. Lombardi \etal~\cite{lombardi2018deep} pioneered joint modeling of shape and appearance using VAEs, while works like Bagautdinov \etal~\cite{bagautdinov2018modeling} and Ranjan \etal~\cite{ranjan2018generating} employed mesh convolutions for detailed geometry capture. FLAME~\cite{li2017learning} incorporated linear blend skinning for jaw and neck movements but had difficulty conveying subtle expressions.
Recent advances have focused on improving rendering quality and generalization. The Pixel Codec Avatar (PiCA)~\cite{ma2021pixel} introduced dynamic texture resolution through pixel-based decoding, departing from traditional fixed texture maps~\cite{lombardi2018deep} and vertex-based representations~\cite{zhou2019dense}. However, PiCA's subject-specific nature and single-mesh representation limit its scalability and hair modeling capabilities. Universal models like LatentAvatar~\cite{xu2023latentavatar} and URAvatar~\cite{li2024uravatar} have attempted to address generalization across identities, but often struggle with preserving person-specific details and handling large deformations, particularly in hair regions. Cao \etal~\cite{cao2022authentic} proposed a shared expression space across identities, but accurate guide meshes remained challenging, affecting reconstruction quality.
Our work addresses these limitations by combining PiCA's efficient and accurate pixel-based rendering with a Universal Prior Model for cross-identity generalization. Crucially, we introduce a layered representation that separates face and hair components, enabling better alignment and optimization compared to single-mesh approaches while maintaining high visual fidelity across different identities, expressions and poses.

\noindent \textbf{3D Hair Modeling.}
Hair modeling in 3D avatar reconstruction has been explored through various approaches. Traditional strand-based methods, whether using multiview stereo~\cite{paris2008hair, luo2012multi} or single-view inference~\cite{chai2012single, zhou2018hairnet, zheng2023hairstep,wu2024monohair}, focus on explicit strand geometry recovery. While these methods can achieve high geometric accuracy, they are often computationally intensive and impractical for mobile VR applications. Alternative approaches have explored different representations for hair modeling. HeadCraft~\cite{sevastopolsky2025headcraft} combines parametric head models with StyleGAN-generated displacement maps for animation control and detail preservation. Volumetric methods like Neural Volumes~\cite{lombardi2019neural} and MVP~\cite{lombardi2021mixture} have demonstrated impressive results in hair rendering, with HVH~\cite{wang2022hvh} and NeuWigs~\cite{wang2023neuwigs} further improving hair animation through layered modeling. However, these person-specific models often struggle with generalization to novel identities.
While recent works have attempted to address generalization through pixel-aligned information~\cite{raj2021pixel} or cross-identity hypernetworks~\cite{cao2022authentic}, they either face challenges with complex geometry or depend heavily on precise head mesh tracking. Our method focuses on a universal compositional representation that separately models face and hair components using efficient mesh-based representations, enabling real-time rendering while maintaining visual quality across diverse hairstyles and identities.

\noindent \textbf{Compositional Avatar Representation.}
Compositional modeling has emerged as a promising direction for improving the quality and controllability of 3D avatars.
% Recent works have explored this approach for different aspects of human representation. 
For face and accessories, MEGANE~\cite{li2023megane} demonstrated the benefits of compositional modeling by combining surface geometry and volumetric representation for eyeglasses, enabling accurate geometric and photometric interactions with faces. DELTA~\cite{feng2023learning} proposed a hybrid explicit-implicit representation to disentangle face and hair components, but primarily focused on static reconstruction and hairstyle transfer.
For head avatars, RGCA~\cite{saito2024relightable} and URAvatar~\cite{li2024uravatar} have shown the advantages of separately modeling head and eye regions for better eye dynamics and relighting effects.
% In the domain of full-body reconstruction, w
Works like GALA~\cite{kim2024gala} and LayGA~\cite{lin2024layga} have introduced layered representations that decompose body and clothing, showing improved results in clothing dynamics and detail preservation. TECA~\cite{zhang2024teca} further extends compositional modeling to text-guided avatar generation.
% , using different representations (meshes and NeRFs) for different components to better handle their distinct structural qualities.
Unlike previous works that treat the head as a single entity or focus on static composition, we introduce a layered representation specifically designed to capture the complex dynamic interactions between face and hair during expressions and pose changes. Our approach uniquely combines compositional modeling with a universal prior model, enabling consistent expression transfer and pose-dependent hair dynamics across different identities.

% \di{Giljoo please help refine and shorten the related work.}
\section{Methods} Our approach starts with generating assets for our captured and tracked datasets (Sec.\ref{sec:dataset}), which support the universal layered prior model (Sec.\ref{sec:upm_layered}), based on a novel mesh-based UPM (Sec.\ref{sec:upm_pica}). Our layered design improves anchor geometry for precise Gaussian renderings (Sec.\ref{sec:urgca}). Loss functions and training details are outlined in Sec.~\ref{sec:training_losses}.
\begin{figure}[t]
% \vspace{-15pt}
\centering
\includegraphics[width=1\linewidth]{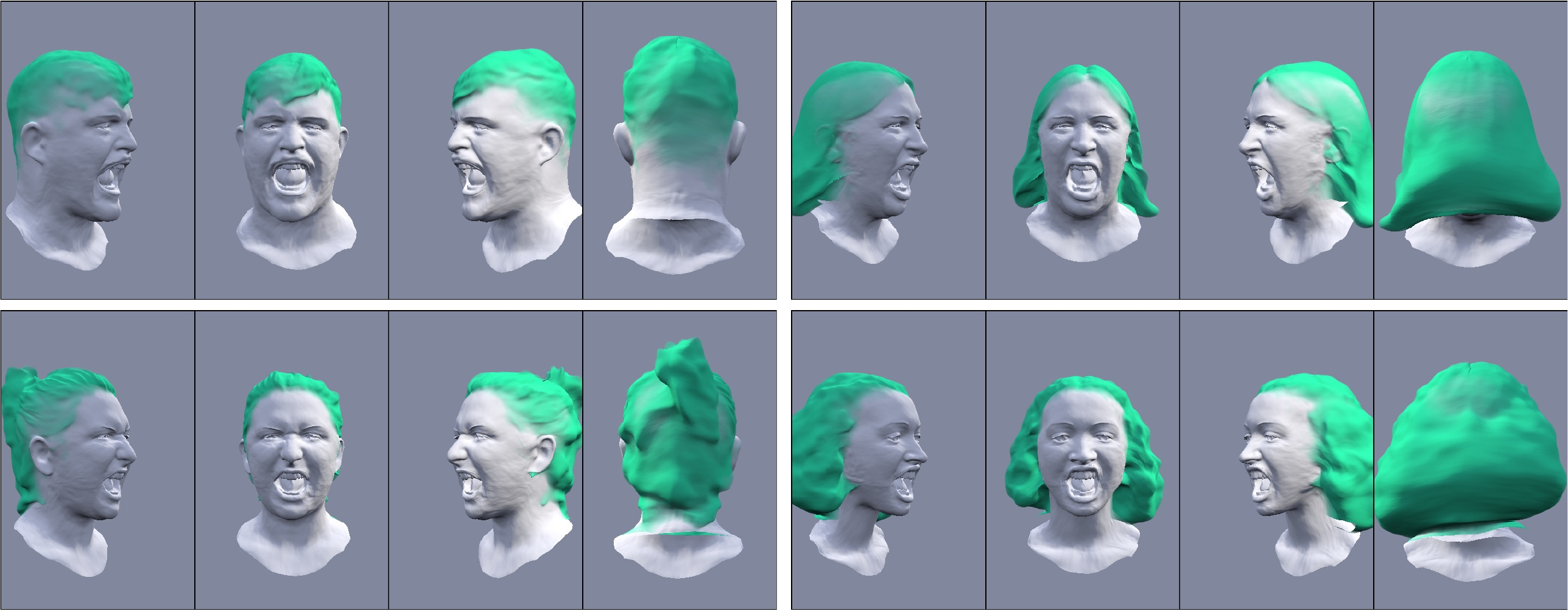}
\vspace{-15pt}
\caption{
\textbf{Dehaired Head and Hair Geometries.} Our method precisely disentangles dehaired head from hair for different users.}
\label{fig:dehair}
\vspace{-10pt}
\end{figure}

\subsection{Dataset and Assets}
\label{sec:dataset}
% We employ the multi-view capture system described by Cao \etal~\cite{cao2022authentic} to record the facial performances of 76 participants. Each participant's dataset comprises approximately 14,000 frames captured from 110 distinct camera angles. Notably, 12 of these participants are bald. Initially, we tracked the geometries in the datasets of these 12 bald participants and constructed a Principal Component Analysis (PCA) model~\cite{abdi2010principal} based on these geometries. This PCA model serves as a re-projection loss during the tracking of geometries in the datasets of other participants, ensuring precise supervision of the bald head geometry by using the dehaired mesh as the ground truth. Additionally, we deform a hair template with 20k vertices to track each participant's hair on top of these dehaired geometries. Examples of dehaired head and hair geometries are illustrated in~Fig.~\ref{fig:dehair}.

We use the multi-view capture system from Cao \etal~\cite{cao2022authentic} to record facial performances of 76 identities 
%which comprises approximately of 14,000 frames 
captured from 110 distinct cameras.
To learn 2D hair segmentation textures for each identity, we add predicted segmentation masks from HRNet~\cite{Wang2019DeepHR}, trained on our in-house dataset. 
Notably, 5 identities are bald.
Starting with these bald individuals, we iteratively build a linear deformable model~\cite{Blanz1999AMM} of bald head geometry, gradually expanding by “dehairing” the next participant with least amount of hair until covering all 76 identities, see Fig.~\ref{fig:dehair}.
We learn the linear deformable model by using Expectation Maximization (EM) for factor analysis\cite{Ghahramani1996TheEA}, following Torresani~\cite{NIPS2003_8db92642}, and find that adding Laplacian smoothness loss to the M-step can help regularize shapes yielding better results compared to vanilla PCA~\cite{abdi2010principal}.
Dehairing is performed by computing the expected values of latent variables similar to the E-step by only using the observed data (excluding hair-covered areas) which in turn is used to infer the hidden bald geometry which we use to inpaint the hair regions, stitched using~\cite{SorkineHornung2007AsrigidaspossibleSM}.
% 
% Examples of dehaired head and hair geometries are illustrated in~Fig.~\ref{fig:dehair}.

%\cc{Di could you fill the numbers here?}~\di{done.} 
% \di{BTW, we had this dataset section but moved to supp due to space limit.}

\begin{figure*}[t]
% \vspace{-15pt}
\centering
\includegraphics[width=1\linewidth]{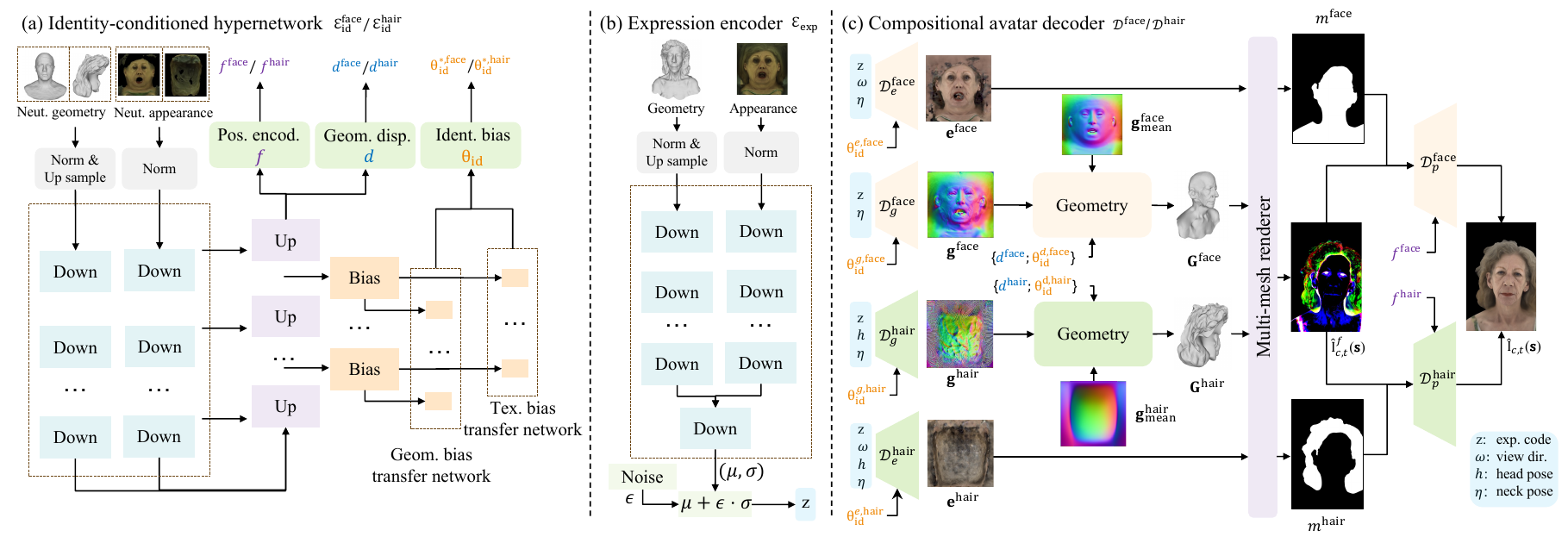}
\vspace{-20pt}
\caption{
\textbf{Overview of LUCAS.} (a) Our identity-conditioned hypernetwork $\mathcal{E}_\text{id}^\text{face}$/$\mathcal{E}_\text{id}^\text{hair}$ generates identity-specific features $\{f, d\}$ and untied biases $\Theta_\text{id}$ from neutral geometry and appearance data. (b) The expression encoder $\mathcal{E}_\text{exp}$ learns a unified expression code space that enables consistent expression transfer across identities. (c) Given expression code $z$, view direction $\omega$, and poses $\{h, \eta\}$, our compositional avatar decoder $\mathcal{D}^\text{face}$/$\mathcal{D}^\text{hair}$ produces separate geometry and appearance maps for face and hair. These are combined with mean geometry and geometry displacement for multi-mesh rendering, followed by separate pixel decoders for the final avatar image generation.
}
\label{fig:flowchart}
% \vspace{-2mm}
\end{figure*}

\subsection{Universal Prior Model for Pixel Codec Avatars}
\label{sec:upm_pica}

Pixel Codec Avatars (PiCA)~\cite{ma2021pixel} offers precise mesh tracking and real-time rendering but limited to personalized models. 
%
%Inspired by~\cite{cao2022authentic}, we propose a Universal Prior Model (UPM) for PiCA. 
Inspired by~\cite{cao2022authentic}, we extend Personalized PiCA with cross-identity capacity, powered by a Universal Prior Model (UPM). We call the new model as \textbf{uPiCA}.
% we propose a Universal Prior Model (UPM) for PiCA.
%Similarly, we employ an identity-conditioned hypernetwork~\cite{ha2016hypernetworks} to generate person-specific avatars. The hypernetwork $\mathcal{E}_\text{id}$ takes identity features as inputs and produces a subset of person-specific model weights.
Similarly, uPiCA adopts a Variational AutoEncoder (VAE)~\cite{kingma2013auto} architecture with an expression encoder, and an avatar decoder. Besides, an identity-conditioned hypernetwork~\cite{ha2016hypernetworks} is added to generate person-specific avatars.
%uPiCA is composed of a expression encoder, an avatar decoder, and an identity-conditioned hypernetwork~\cite{ha2016hypernetworks} to generate person-specific avatars.

\noindent \textbf{Identity-conditioned hypernetwork}
%To enable the extraction of person-specific details, 
$\mathcal{E}_\text{id}$ takes a neutral texture map $\textbf{T}_{\text{neu}}$ and a neural geometry image (mapping vertex position to texture UV space), $\textbf{G}_{\text{neu}}$, and generates bias maps $\Theta_{\text{id}}$ for each level of the avatar decoder $\mathcal{D}$ via a set of skip connections. $\mathcal{E}_\text{id}$ also generates a per-identity positional encoding $f$ for pixel decoder and per-identity geometry displacement $d$ for geometry decoder:
%, similar to a U-Net architecture \cite{ronneberger2015u}:
\begin{equation}
f, d, \Theta_{\text{id}} = \mathcal{E}_\text{id}(\textbf{T}_{\text{neu}}, \textbf{G}_{\text{neu}}; \Phi_{\text{id}}). 
\label{hypernet}
\end{equation}
$\Phi_{\text{id}}$ is the trainable parameters for the identity encoder. 
%
% The decoder is also conditioned on view direction $\omega$, and neck pose $\eta$, used for rendering, to allow explicit control over view- and pose-dependent appearance changes.~\rongyu{Should been moved to decoder.}

\noindent \textbf{Expression encoder.}  
The expression code \( z \) are generated by the expression encoder \( \mathcal{E}_{\text{exp}} \), which takes the differences between the current and neutral geometry and texture maps as input:  
\(
\Delta \textbf{G}_{\text{exp}} = \textbf{G}_{\text{exp}} - \textbf{G}_{\text{neu}},  
\Delta \textbf{T}_{\text{exp}} = \textbf{T}_{\text{exp}} - \textbf{T}_{\text{neu}},
\)  
where \( \textbf{G}_{\text{exp}} \) and \( \textbf{T}_{\text{exp}} \) are the current geometry and texture maps, respectively. \( z \in \mathbb{R}^{16 \times 4 \times 4} \) is defined as:
\begin{equation}
%\begin{split}
    z =  \mathcal{N}(\mu, \sigma); \ \mu, \sigma = \mathcal{E}_{\text{exp}}(\Delta \textbf{T}_{\text{exp}}, \Delta \textbf{G}_{\text{exp}}; \Phi_{\text{exp}}),
%\end{split}
\end{equation}
where 
%\( \mathcal{N}(0, 1) \) is the unit normal distribution, and
\( \Phi_{\text{exp}} \) are the trainable parameters of $\mathcal{E}_{\text{exp}}$. Since the model is trained end-to-end on multi-identity data, the same expression code can be reused across different identities for driving, ensuring consistent expression transfer.

\noindent \textbf{Avatar decoder.} 
We use a set of multiview images $\mathbf{I}_{c,t}$ (\ie, images from camera $c$ at frame $t$) with calibrated intrinsics $\mathbf{K}_c$ and extrinsics $\mathbf{R}_c \mid \mathbf{t}_{c}$.
% \rongyu{Extrinsics should be $\mathbf{R} \mid \mathbf{t}_{c}$, without frame $t$?} ~\di{fixed.}
%
% We compute the camera viewing direction as $\omega = \mathbf{R}_c ^\top \mathbf{t}_{c} $ and transform this vector into an $16 \times 8 \times 8$ grid through a linear layer. 
To condition the decoder on the view direction, we compute $\omega = \mathbf{R}_c^\top \mathbf{t}_c$ (approximating the viewing direction based on a head-centered coordinate system). This vector is transformed into a $16 \times 8 \times 8$ grid via a linear layer.
Additionally, we enhance the geometry to encompass the shoulder region and use linear blend skinning to model neck pose $\eta \in \mathbb{R}^{6}$. We use a similar decoder architecture $\mathcal{D}$ as PiCA, which consists of an appearance decoder $\mathcal{D}_e$, a geometry decoder $\mathcal{D}_g$, and a pixel decoder $\mathcal{D}_p$. 
Noted that the outputs of geometry and appearance decoder are both expression-dependent.
The geometry decoder takes the latent code $z$ and neck pose $\eta$ as input and decodes a head-centered 3D dense position map.
The appearance decoder uses the latent code $z$, viewing direction $\omega$, and neck pose $\eta$ to decode a low-resolution, view-dependent map of local appearance codes:
\begin{equation}
\begin{split}
    % \mathbf{g} & = \mathcal{D}_g\left(z, \eta; \Theta_{\text{id}}^g, \Phi_g\right), \\
    % \mathbf{e} & = \mathcal{D}_e\left(z, \omega, \eta; \Theta_{\text{id}}^e, \Phi_e\right),
    \mathbf{g}  = \mathcal{D}_g\left(z, \eta; \Theta_{\text{id}}^g, \Phi_g\right); \quad
    \mathbf{e} = \mathcal{D}_e\left(z, \omega, \eta; \Theta_{\text{id}}^e, \Phi_e\right),
\end{split}
\label{ge}
\end{equation}
where $\mathbf{g} \in \mathbb{R}^{256 \times 256 \times 3}$ is a map of geometry displacement, and $\mathbf{e} \in \mathbb{R}^{256 \times 256 \times 4}$ is a map of appearance codes. $\Theta_{\text{id}}^*$ are identity-specific biases from Eq.~\ref{hypernet} and are related to the corresponding decoders $\mathcal{D}_*$. $\Phi_*$ are their corresponding network training parameters.
We define the final geometry as
$\mathbf{G} = \mathbf{g}_\text{mean} + d + \mathbf{g}$,
% ~\rongyu{Shall we also bold $d$?} ~\di{I think no.}
where we apply a Laplacian preconditioning~\cite{Nicolet2021Large} to the gradients of mean geometry \(\mathbf{g}_\text{mean} \) to bias gradient steps towards smooth solutions. 
\( d \) and \( \mathbf{g} \) are the per-identity and expression-dependent geometry displacement, respectively.
The final geometry \( \mathbf{G} \) is sampled at each vertex’s UV coordinates to produce a mesh for rasterization with $\mathbf{e}$. 
Rasterization assigns to a pixel at screen position $\textbf{s}$ its corresponding UV coordinates $\textbf{u}$ and head-centered $xyz$ coordinates $\textbf{x}$, and produces the feature image $\hat {\mathbf{I}}_{c,t}^f(\mathbf{s})$. 
The pixel decoder further decodes the color at each pixel to produce the rendered image through:
\begin{equation}
        \hat {\mathbf{I}}_{c,t}(\mathbf{s}) = \mathcal{D}_p\left(\hat {\mathbf{I}}_{c,t}^f(\mathbf{s}), f, \textbf{x}, \textbf{u}; \Phi_p\right),   
\label{p}
\end{equation}
where $\Phi_p$ are the training parameters of $\mathcal{D}_p$. $f$ is the positional encoding from Eq~\ref{hypernet}. 
Note that $\mathcal{D}_p$ uses shared weights across subjects, avoiding identity-specific biases from the hypernetwork. Appearance variations are effectively captured by the feature inputs, enhancing network efficiency for runtime deployment and eliminating the need to manage multiple shaders for different users.

\subsection{Universal Layered Prior Model}
\label{sec:upm_layered}

To enable a universal prior model for compositional face and hair avatars across identities, we extend uPiCA to a layered approach, as shown in Fig.~\ref{fig:flowchart}. 
In this model, we employ two parallel hypernetworks for face and hair, \ie, $\mathcal{E}_\text{id}^{\text{face}}$ and $\mathcal{E}_\text{id}^{\text{hair}}$.
%to generate untiled bias maps for the face and hair decoders. 
%
This separation allows the model to capture intricate details, such as hair deformation due to head movements or facial expressions.
% , while maintaining the geometric coherence between the face and hair regions. 
%
We employ a unified expression encoder that extracts shared features from the tracked data, enabling synchronized control of both face and hair deformations through a common expression space.
The encoded information is then passed in parallel to two independent decoders, $\mathcal{D}^\text{face}$ and $\mathcal{D}^\text{hair}$, allowing each part to adapt to its unique geometry and appearance.

\noindent \textbf{Compositional avatar decoder.}
We use the same decoder architecture as uPiCA from Sec.~\ref{sec:upm_pica} for the face decoders, and denote them as $\mathcal{D}_g^\text{face}$, $\mathcal{D}_e^\text{face}$, and $\mathcal{D}_p^\text{face}$. For the hair geometry decoder $\mathcal{D}_g^\text{hair}$, we use both the head pose \( h \) and neck pose \( \eta \) as inputs since these factors influence hair movement. 
Additionally, we include the latent code \( z \) as the input of $\mathcal{D}_g^\text{hair}$, as our experiments reveal that hair deforms with certain facial expressions, such as frowning.
% ~\rongyu{Need to make sure we have ablation study for this claim}.~\di{yes we have.}
This behavior arises because the skin beneath the hair shifts with facial movements, causing the hair to adjust accordingly.
For the hair appearance decoder $\mathcal{D}_e^\text{hair}$, we take all inputs of $\mathcal{D}_g^\text{hair}$ along with the view direction \( \omega \) to account for view-dependent appearance variations, ensuring that both the geometry and texture adapt seamlessly across viewing angles. The layered hair decoders are formulated as:
\begin{equation}
\begin{split}
    \mathbf{g}^{\text{hair}} & = \mathcal{D}_g^{\text{hair}}(z, \eta, h; \Theta^{g,\text{hair}}_\text{id}, \Phi_g^{\text{hair}}); \\
    \mathbf{e}^{\text{hair}} & = \mathcal{D}_e^{\text{hair}}(z, \omega, \eta, h; \Theta^{e,\text{hair}}_\text{id}, \Phi_e^{\text{hair}}),
\end{split}
\label{ge_hair}
\end{equation}
where \( \mathbf{g}^{\text{hair}} \in \mathbb{R}^{256 \times 256 \times 3} \) and \( \mathbf{e}^{\text{hair}} \in \mathbb{R}^{256 \times 256 \times 4} \) are the position and texture map of the hair mesh, respectively. 
% The hair geometry decoder extracts a mesh with 20k vertices, which captures detailed hair shapes and ensures smooth integration with the face geometry.

\noindent \textbf{Multi-mesh joint rendering.}
After decoding the face and hair components, we obtain two geometry maps: \( \mathbf{G}^\text{face} \) and \( \mathbf{G}^{\text{hair}} \). These maps are concatenated and jointly processed with the texture maps \( \mathbf{e}^\text{face} \) and \( \mathbf{e}^{\text{hair}} \) using a differentiable renderer~\cite{Pidhorskyi2024RasterizedEG} to produce a unified feature vector for the entire screen image. We further apply the face and hair mask \( m^{\text{face}} \) and \( m^{\text{hair}} \) to the rendered feature map and feed the masked feature images $\hat {\mathbf{I}}_{c,t}^{f,\text{face}}(\mathbf{s})$ and $\hat {\mathbf{I}}_{c,t}^{f,\text{hair}}(\mathbf{s})$, along with their corresponding \( \mathbf{x} \) and \( \mathbf{u} \) into separate pixel decoders $\mathcal{D}_p^\text{face}$ and $\mathcal{D}_p^\text{hair}$. The final rendered image is given by:
\begin{equation}
\begin{aligned}
    \hat{\mathbf{I}}_{c,t}^\text{l}(\mathbf{s})
    &= \mathcal{D}_p^\text{face}(\hat {\mathbf{I}}_{c,t}^{f,\text{face}}(\mathbf{s}), \mathbf{x}, \mathbf{u}; \Phi_p^\text{face}) \odot m^{\text{face}} \\
    &+ \mathcal{D}_p^{\text{hair}}(\hat {\mathbf{I}}_{c,t}^{f,\text{hair}}(\mathbf{s}), \mathbf{x}, \mathbf{u}; \Phi_p^{\text{hair}}) \odot m^{\text{hair}}.
\end{aligned}
\end{equation}
% where \( m^{\text{face}} \) and \( m^{\text{hair}} \) are the masks for the face and hair regions, respectively, obtained from the multi-mesh rasterization. 
% This layered design enables precise alignment between the face and hair components, ensuring high-fidelity reconstruction while supporting complex hairstyles and dynamic movements.
\subsection{Layered Meshes for Gaussian Rendering}
\label{sec:urgca}
We show that our layered mesh design improves the anchor geometry for precise and visually appealing Gaussian renderings. 
Building on prior works~\cite{li2024uravatar,cao2022authentic}, we parameterize and anchor Gaussians on the vertices of our layered PiCA guide mesh. 
We employ parallel face and hair branches for the Gaussian hypernetwork and decoder, which share the same architecture. For simplicity, we denote the hypernetwork and decoder for each branch as $\mathcal{E}_\text{id}^\text{gs}$ and $\mathcal{D}^\text{gs}$. The hypernetwork is formulated as:
\begin{equation}
d^c_\text{mean}, \Theta_\text{id}^{\text{gs}} = \mathcal{E}_\text{id}^{\text{gs}}(\textbf{T}_{\text{neu}}, \textbf{G}_{\text{neu}}; \Phi_{\text{id}}^\text{gs}),
\label{e_id}
\end{equation}
where $d^c_\text{mean}$ represents the mean color attribute from neutral appearance data, and $\Theta_\text{id}^{\text{gs}}$ is the identity-specific bias map.
We denote the vertex positions of the layered PiCA guide mesh as \( \{ \hat{t}_k \}_{k=1}^M \), which serve as anchors for the Gaussians. The Gaussian decoder \( \mathcal{D} ^{\text{gs}} \) takes the expression code \( z \) and neck pose \( \eta \) as input, and is conditioned on the identity untied bias map \( \Theta_{\text{id}}^{\text{gs}} \). It outputs the following attributes:
\begin{equation}
\left\{ \delta t_k, q_k, s_k, d_k^c, o_k \right\}_{k=1}^M = 
\mathcal{D}^{\text{gs}}\left(z, \eta; \Theta_{\text{id}}^{\text{gs}}, \Phi^{\text{gs}}\right), \tag{9}
\end{equation}
where \( \delta t_k \) is the position delta, \( q_k \) is the rotation quaternion, \( s_k \) is the scale, \( d_k^c \) is the color attribute, and \( o_k \) is the opacity. 
% Here, \( \Phi^{\text{gs}} \) represents the learnable parameters of the Gaussian decoder. 
The final Gaussian positions are computed as $t_k = \hat{t}_k + \delta t_k$, and colors as $d^c_\text{mean} + d^c_k$ for rendering.

\subsection{Training and Losses}
\label{sec:training_losses}

We jointly optimize all the trainable network parameters $\Phi$ using a total loss \( \mathcal{L}_{\text{total}} \) consisting of:
\begin{equation}
\mathcal{L}_{\text{total}} = \lambda_{\text{pica}} \mathcal{L}_{\text{pica}} + \lambda_{\text{gs}} \mathcal{L}_{\text{gs}} + \lambda_{\text{dehair}} \mathcal{L}_{\text{dehair}},
\end{equation}
where \( \mathcal{L}_{\text{pica}} \) and \( \mathcal{L}_{\text{gs}} \) are the PiCA reconstruction and Gaussian losses, respectively, and \( \mathcal{L}_{\text{dehair}} \) is the dehairing loss. \( \lambda_* \)  are their corresponding loss weights. For the dehairing loss \( \mathcal{L}_{\text{dehair}} \), a large initial weight is applied with a decay during training to accelerate the convergence of bald geometry, ensuring accurate dehaired geometry without interference from the hair mesh. The dehaired avatar serves as the foundation for adding a hair layer, allowing joint optimization of both face and hair with precise alignment.

\noindent \textbf{PiCA reconstruction loss.} We extend the original PiCA losses in \cite{ma2021pixel} to a layered reconstruction loss, defined as:
\begin{equation}
\begin{split}
\mathcal{L}_{\text{pica}} & = \lambda_I \mathcal{L}_I + \lambda_D \mathcal{L}_D + \lambda_N \mathcal{L}_N + \lambda_M \mathcal{L}_M \\
& + \lambda_S \mathcal{L}_S + \lambda_{\text{KL}} \mathcal{L}_{\text{KL}} + \lambda_{\text{seg}} \mathcal{L}_{\text{seg}}.
\end{split}
\end{equation}
Here, \( \mathcal{L}_I \), \( \mathcal{L}_D \), \( \mathcal{L}_N \), and \( \mathcal{L}_{\text{KL}} \) correspond to photometric, depth, normal, and KL divergence losses, respectively, as defined in the original PiCA paper~\cite{ma2021pixel}. The photometric loss \( \mathcal{L}_I \) measures the $L_1$ difference between predicted and ground truth images. 
The mesh tracking loss \( \mathcal{L}_M \) handles hair and face meshes separately by leveraging the tracked hair mesh and the dehaired geometry from avatar dehairing. Smoothness terms \( \mathcal{L}_S \), including Laplacian and general smoothness regularization, are applied independently to both hair and face meshes to prevent artifacts caused by noisy depth inputs, incomplete depth supervision, or stochastic gradient descent noise. The segmentation loss \( \mathcal{L}_{\text{seg}} \) ensures accurate reconstruction of hair regions, particularly thin strands along the sides of the head, preventing them from blending into the face mesh. Notably, we refine the hair segmentation mask through erosion and dilation, creating a mask that extends beyond exact boundaries to account for inaccuracies, and weights \( \mathcal{L}_{\text{seg}} \).

\begin{figure*}[t]
% \vspace{-15pt}
\centering
\includegraphics[width=1\linewidth]{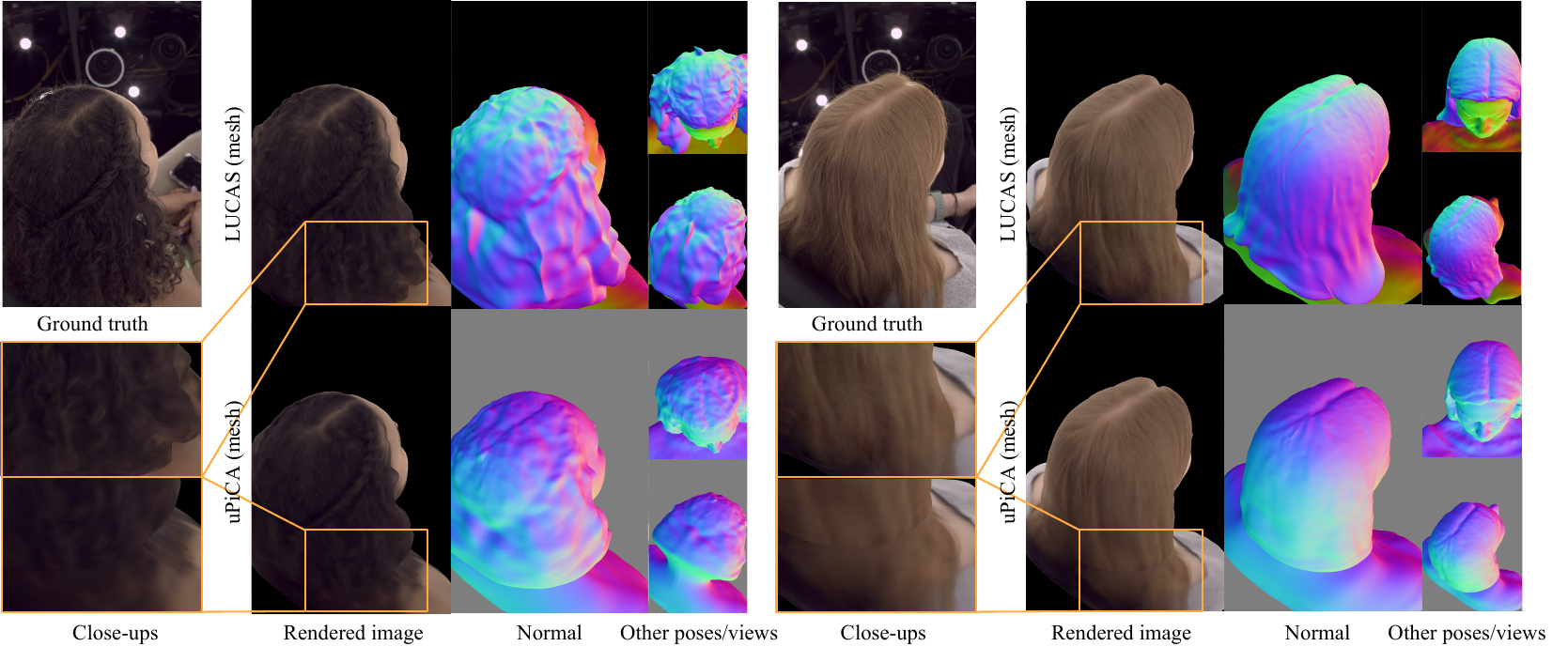}
\vspace{-20pt}
\caption{
\textbf{Qualitative comparison (mesh).} Our layered representation enables better reconstruction of long hair compared to uPiCA's single-mesh approach. While uPiCA struggles with hair-shoulder intersections and loses hair tail details during head movement, our method maintains clean geometry with accurate hair shape and positioning across different head poses.}
\label{fig:comparison_mesh}
% \vspace{-2mm}
\end{figure*}

\noindent \textbf{Gaussian loss.} The parameters of the Gaussian branch are optimized using the following loss:
\vspace{-5pt}
\begin{equation}
\mathcal{L}_{\text{gs}} = \lambda_\text{render} \mathcal{L}_\text{render} + \lambda_\text{scale} \mathcal{L}_\text{scale} + \lambda_\Delta \mathcal{L}_\Delta. 
\end{equation}
The Gaussian render loss \( \mathcal{L}_{\text{render}} \) applies the $L_1$ loss on the rendered image, following the original 3DGS paper~\cite{kerbl20233d}. To regularize the scale of the Gaussian primitives, we define a pre-clamped scale regularization loss as:
\begin{equation}
\begin{split}
\mathcal{L}_\text{scale} & = \frac{1}{M} \sum_{k=1}^{M} 
\left( 
\frac{1}{\max(r_\text{min}, s_k)} \cdot \mathbb{I}(s_k < r_\text{min}) \right. \\
& \quad + \left. \left( \max(0, s_k - r_\text{max}) \right)^2 
\right),
\end{split}
\end{equation}
% \begin{equation}
% \mathcal{L}_\text{scale} = \frac{1}{N} \sum_{k=1}^{N} 
% \left( 
% \frac{\mathbb{I}(s_k < r_\text{min}) }{\max(r_\text{min}, s_k)} + \left( \max(0, s_k - r_\text{max}) \right)^2
% \right),
% \end{equation}
where \( s_k \) is the scale value of the \( k \)-th Gaussian primitive along any axis, and \( M \) is the total number of primitives. The variables \( r_\text{min} \) and \( r_\text{max} \) represent the lower and upper bounds of the primitive scale, set to 0.1 and 5.0 in our experiments. Note that the regularization loss is computed on the original, unclamped scales \( s_k \) to penalize deviations effectively. We clamp the primitive scale values to the range \([r_\text{min}, r_\text{max}]\) before passing them to the Gaussian renderer, ensuring the rendered Gaussians remain within a controlled range. \( \mathbb{I}(\cdot) \) denotes the indicator function, which equals 1 if the condition is true and 0 otherwise. This formulation ensures stable optimization by keeping the Gaussian scales within appropriate bounds.
Moreover, we apply a delta position loss \( \mathcal{L}_\Delta \) to both the hair and face Gaussians to prevent them from drifting too far from their guide mesh. This loss ensures that hair Gaussians stay within the hair area, and Gaussians on the bald regions of the face mesh do not migrate into the hair region. Specifically, the loss penalizes position deviations as follows:
\begin{equation}
\mathcal{L}_\Delta = \mathbb{E}\left[ \left( \delta t^{\text{hair}} \right)^2 \right] 
+ \mathbb{E}\left[ \left( \delta t^{\text{face}} \odot (1 - m^{\text{face}}) \right)^2 \right],
\end{equation}
where \( \delta t^{\text{hair}} \) and \( \delta t^{\text{face}} \) represent the position deltas of the hair and face Gaussians, respectively, and \( m^{\text{face}} \) is the face mask used to ensure that the delta loss is only applied to the bald head region. More training details are given in \textit{suppl.}

% This formulation ensures that the hair Gaussians remain confined to their designated area, while the Gaussians on the bald regions of the head do not encroach into the hair region, maintaining the separation between face and hair areas throughout training.

% The relative weights are:
% $\lambda_{\text{geo}} = 1, \quad \lambda_{l1} = 10, \quad \lambda_{\text{ssim}} = 0.2, \quad \lambda_s = \lambda_{c-} = \lambda_{\text{es}} = 1.0 \times 10^{-2}, \quad \lambda_{\text{eo}} = \lambda_{\text{ev}} = 1.0 \times 10^{-4}, \quad \lambda_{\text{kl}} = 2.0 \times 10^{-3}.$

\begin{figure*}[t]
% \vspace{-15pt}
\centering
\includegraphics[width=1\linewidth]{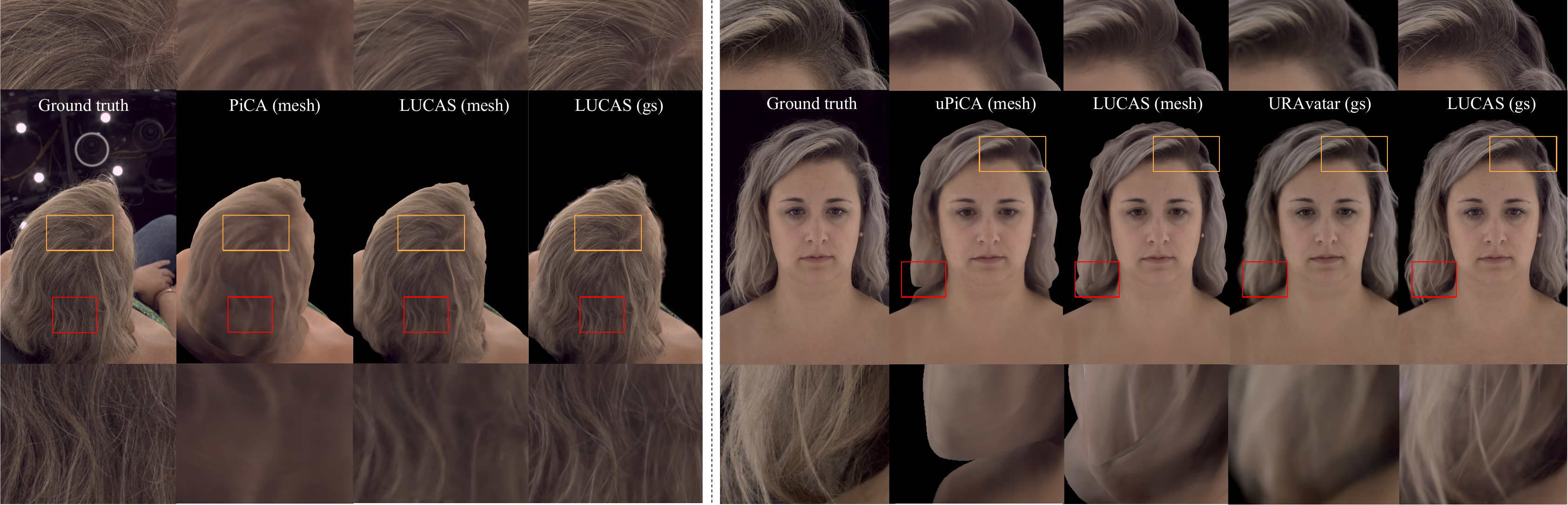}
\vspace{-15pt}
\caption{
\textbf{Qualitative comparison.} Left: Comparison with personalized models shows our method achieves more precise hair reconstruction than PiCA's mesh results. Right: In comparison with universal models, while uPiCA exhibits artifacts such as hair growing from shoulders, our LUCAS (mesh) achieves cohesive reconstruction. When rendered with Gaussian splatting, LUCAS (gs) demonstrates superior detail preservation compared to URAvatar, particularly in complex hairstyles.}
\label{fig:comparison}
\vspace{-5pt}
\end{figure*}

\section{Experiments}

\textbf{Evaluation protocols.} We adopt three widely-used metrics for evaluation: Peak Signal-to-Noise Ratio (PSNR), Structural Similarity Index Measure (SSIM)~\cite{wang2004image}, and Learned Perceptual Image Patch Similarity (LPIPS)~\cite{zhang2018unreasonable}. We restrict the evaluation to the foreground regions, as defined by masks derived from the reconstructed geometry.
% For a fair comparison with prior works, all methods are trained and evaluated under a multi-identity setting, unless explicitly stated otherwise.
% For ablation studies, we train with 76 identities and test on 1) unseen segments from a train subject, 2) unseen segments from an unseen subject, and 3) an unseen illumination from a train subject, which orthogonally eval- uate the generalization of our model to novel poses, identi- ties, and illuminations.

\noindent \textbf{Baselines.} 
For mesh-based methods, we primarily compare with Universal PiCA (uPiCA), which extends Pixel Codec Avatars (PiCA)\cite{ma2021pixel} by incorporating our proposed Universal Prior Model (UPM), as detailed in Sec.~\ref{sec:upm_pica}. Additionally, we perform per-identity comparisons with PiCA to evaluate personalized reconstruction performance.
For Gaussian-based methods, our main comparison is with URAvatar~\cite{li2024uravatar}, benchmarking our model’s ability to capture fine-grained visual details with Gaussian splatting.

% We also equipped uPiCA with gaussian branch, named UGS-PiCA for comparison.

% \subsection{Qualitative results}
% Fig.~\ref{cover_image} shows that our universal layered avatars generalize to novel identities, views, poses, expressions, and hairstyles. As our avatars share the same global latent space, we illustrate that our universal avatars can be driven consistently across identities under different expressions and poses in real time.

\subsection{Evaluation of the layered representation}
% Comparision with uPiCA

\noindent \textbf{Disentangled representation enhances mesh quality.} A key contribution of our work is the compositional representation of the face and hair as two separate meshes.  This design addresses a fundamental limitation of single-mesh avatars: their constrained UV space allocation, where hair is restricted to a small portion of the UV map while the face dominates.  By allowing separate UV maps for face and hair, our approach enables more accurate representation of complex hairstyles, particularly for long hair. As shown in Fig.~\ref{fig:comparison_mesh}, the comparison across various head poses demonstrates our method's superior capability in reconstructing long hair details. While uPiCA struggles with hair reconstruction, especially during head movement, our approach maintains precise geometry and reduces common artifacts such as hair color bleeding onto shoulders. This improvement becomes particularly evident when the avatar tilts or lowers its head, where our layered representation ensures the hair remains correctly positioned, resulting in visually coherent and realistic renderings.
% A significant reason for this superiority lies in the way UV space is allocated. In single-mesh models, the hair occupies only a small portion of the UV map, with most of the map dedicated to the face. This allocation often results in substantial mesh distortion for the hair, as the limited UV space fails to represent complex hairstyles effectively. In contrast, our layered approach allows each mesh to utilize its own UV map, avoiding distortions that can compromise visual fidelity. 
\begin{figure}[t]
% \vspace{-15pt}
\centering
\includegraphics[width=1\linewidth]{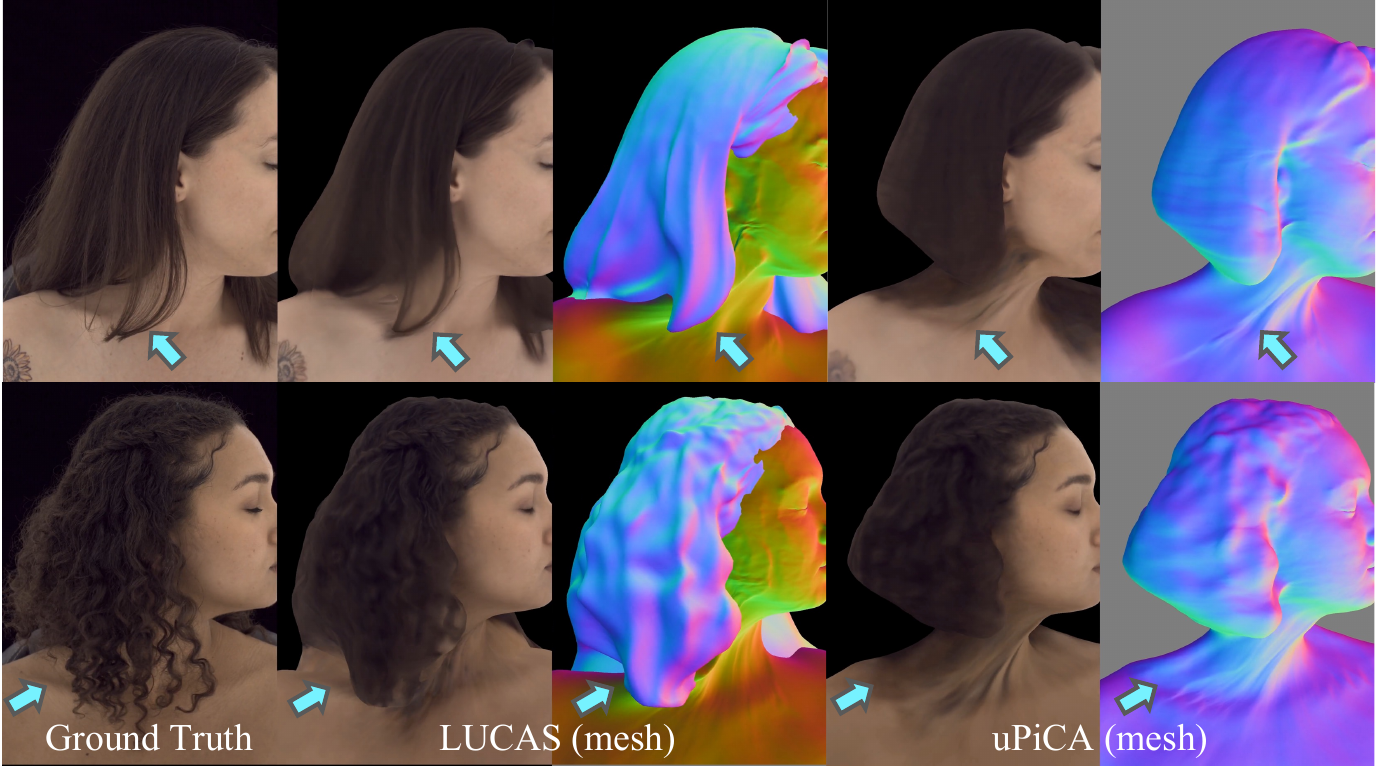}
\vspace{-15pt}
\caption{
\textbf{Comparison on dynamic hair animation.} Our LUCAS mesh tracks hair strand deformation and aligns with head and neck movements, outperforming uPiCA in dynamic scenarios.}
\label{fig:animation}
\vspace{-10pt}
\end{figure}

\noindent \textbf{Improved hair deformation during animation.}
In Fig.~\ref{fig:animation}, we demonstrate the advantage of our method in more dynamic scenarios. The examples show how our LUCAS mesh deforms to match hair strand movement in response to head and neck poses, accurately tracking the motion of long hair. In contrast, uPiCA struggles to adapt the hair strands to the changing head positions, resulting in less natural deformations. This comparison highlights the benefit of our layered approach, which provides better control over hair dynamics and improves realism during animation.

\begin{table}[t]
\centering
\caption{\textbf{Quantitative comparisons} on per-subject ($^{\dagger}$) and cross-subject ($^*$) optimization against the state-of-the-art. 
The top three techniques are highlighted in \textcolor{red!50}{red}, \textcolor{orange!50}{orange}, and \textcolor{yellow!100}{yellow}, respectively.
}
\vspace{-5pt}
\renewcommand\tabcolsep{12pt}
\resizebox{1\linewidth}{!}{
\begin{tabular}{lccc}
\toprule
Method & PSNR $\uparrow$ & SSIM $\uparrow$ & LPIPS $\downarrow$ \\ 
\midrule

$^{\dagger}$PiCA (mesh)~\cite{ma2021pixel} & 32.0512 & 0.8895 & 0.2678   \\

$^{\dagger}$LUCAS (mesh) & \cellcolor{yellow!25}33.5211 & 0.9044 & 0.2479 \\

$^{\dagger}$LUCAS (gs) & \cellcolor{red!25}35.2027 & \cellcolor{red!25}0.9286 & \cellcolor{orange!25}0.2407 \\

\midrule

$^*$uPiCA (mesh) & 32.5623 & 0.8971 & 0.2594 \\

$^*$LUCAS (mesh) & 33.0254 & \cellcolor{yellow!25}0.9073  & 0.2537 \\

$^*$URAvatar (gs)~\cite{li2024uravatar} & 33.1227 & 0.9034 & \cellcolor{yellow!25}0.2464 \\

$^*$LUCAS (gs) & \cellcolor{orange!25}{34.5579} & \cellcolor{orange!25}{0.9201} & \cellcolor{red!25}{0.2394} \\

% $^*$LUCAS (gs) & \textbf{40.0202} & \textbf{0.9711} & \textbf{0.1585} \\

\bottomrule
\end{tabular}}
\vspace{-10pt}
\label{tab:comparison}
\end{table}

\noindent \textbf{Enhanced Gaussian avatars through better meshes.}
The improved mesh structure strengthens the foundation for Gaussian avatars, as the Gaussian splatting process relies heavily on the underlying mesh geometry. While Gaussian splatting can mitigate some errors inherent in single-mesh models, our layered approach further enhances visual fidelity, particularly for intricate hairstyles.
As shown in Fig.~\ref{fig:comparison}, we demonstrate improvements over both personalized and universal models. Compared to PiCA, our approach achieves more detailed hair reconstruction even at the mesh level, with Gaussian splatting further enhancing the visual fidelity. In the universal model comparison, while uPiCA suffers from artifacts like disconnected hair growing from shoulders, LUCAS's mesh representation achieves more cohesive reconstruction. When comparing Gaussian-based methods, LUCAS (gs) demonstrates clear advantages over URAvatar in preserving fine details. These visual improvements are quantitatively validated in Table~\ref{tab:comparison}.

\begin{figure}[t]
% \vspace{-15pt}
\centering
\includegraphics[width=1\linewidth]{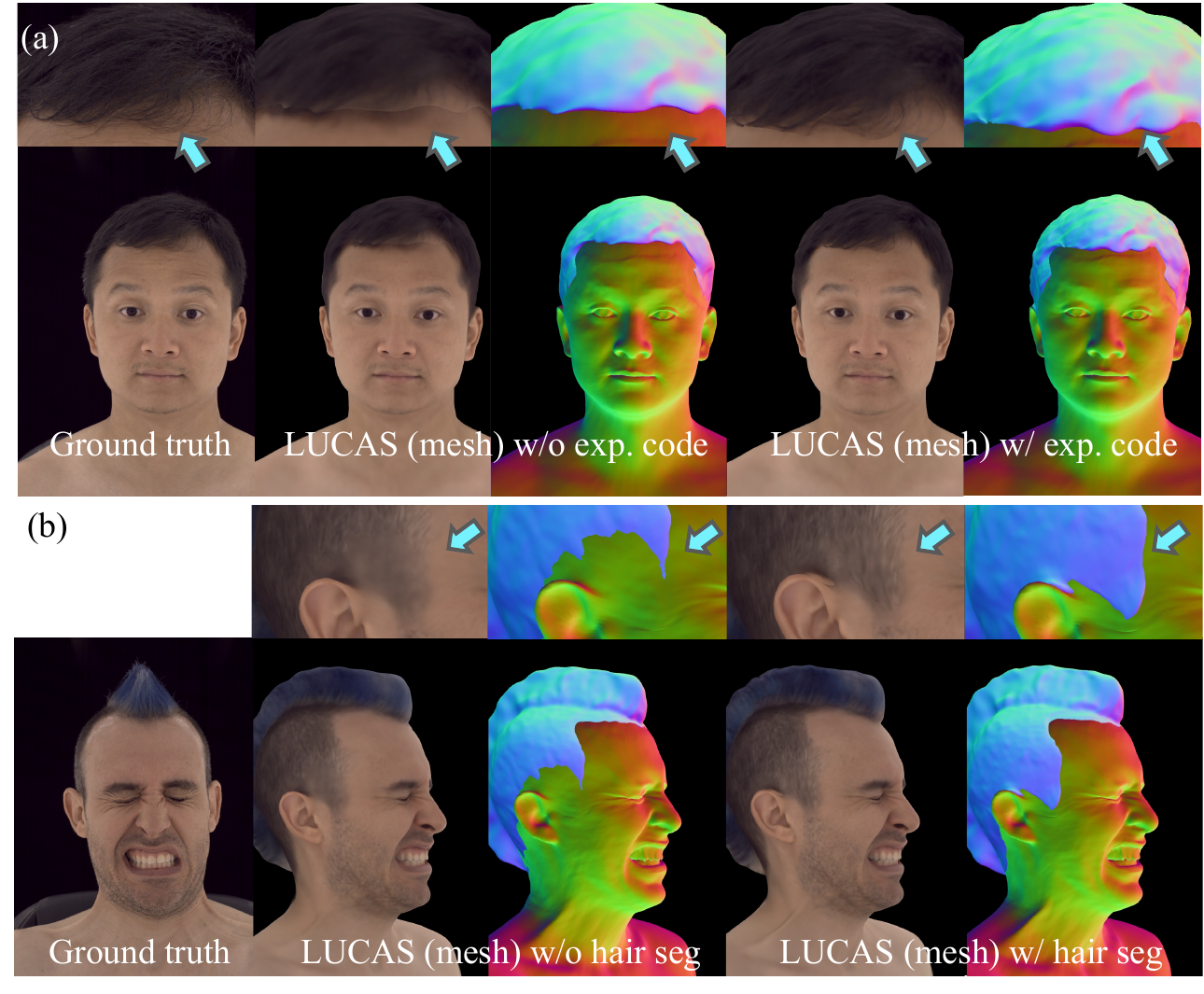}
\vspace{-15pt}
\caption{
\textbf{Ablation study.} (a) Expression code improves face-hair synchronization during expressions. (b) Hair segmentation regularization preserves fine hair details.}
\label{fig:ab_seg_expcode}
% \vspace{-2mm}
\end{figure}

% \noindent \textbf{Compositional avatar representation.} layered head/hair meshes vs. single head mesh.

% \noindent \textbf{Impact of training strategy.} 

\subsection{Ablation study}

\noindent \textbf{Impact of expression code.}
In Fig.~\ref{fig:ab_seg_expcode}(a), we compare results with and without the expression code for hair. Without the expression code, the hair mesh fails to move naturally with facial movements, particularly during expressions like frowning. This observation aligns with the findings in Fig.~\ref{fig:vis_frown}, where a subject looks upward and frowns, the hair should lower slightly toward the eyebrows. Our layered representation enables this natural movement by sharing the same expression code $z$ but decoding it separately for face and hair, allowing each component to deform independently and precisely. This advantage is further validated by the quantitative improvements shown in Table~\ref{tab:ab}.

\noindent \textbf{Impact of segmentation regularization.}
In Fig.~\ref{fig:ab_seg_expcode}(b), we assess the effect of hair segmentation regularization. This component is particularly crucial for reconstructing thin hair, as seen in the example, where the hair on both sides is quite fine. Without segmentation regularization, the mesh struggles to capture these thin strands, resulting in blurred renderings. Adding the segmentation term significantly improves the mesh reconstruction, allowing the fine hair to appear correctly in the final render. Quantitative results are also shown in Table~\ref{tab:ab}.

\begin{table}[t]
\centering
\caption{\textbf{Ablation study} on expression code and hair segmentation regularization, evaluated on both training and unseen subjects.
The top two techniques are highlighted in \textcolor{red!50}{red} and \textcolor{yellow!100}{yellow}, respectively.
}
\vspace{-5pt}
\renewcommand\tabcolsep{2pt}
\resizebox{1\linewidth}{!}{
\begin{tabular}{lcccccc}
\toprule
& \multicolumn{3}{c}{Training subjects} & \multicolumn{3}{c}{Unseen subjects} \\
\cmidrule(lr){2-4} \cmidrule(lr){5-7}
\multirow{1}{*}{Method}  & PSNR $\uparrow$ & SSIM $\uparrow$ & LPIPS $\downarrow$ & PSNR $\uparrow$ & SSIM $\uparrow$ & LPIPS $\downarrow$ \\ 
\midrule
w/o exp. code & \cellcolor{yellow!25}34.1014 & 0.9129 & 0.2498 & \cellcolor{yellow!25}31.9128 & 0.8874 & 0.2601 \\
w/o hair seg & 34.0285 & \cellcolor{yellow!25}0.9140 & \cellcolor{yellow!25}0.2485 & 31.7964 & \cellcolor{red!25}{0.9098} & \cellcolor{yellow!25}0.2554 \\
Full model & \cellcolor{red!25}{34.4981} & \cellcolor{red!25}{0.9189} & \cellcolor{red!25}{0.2402}  & \cellcolor{red!25}{32.5847} & \cellcolor{yellow!25}{0.9087} & \cellcolor{red!25}{0.2496} \\
\bottomrule
\end{tabular}}
\vspace{-5pt}
\label{tab:ab}
\end{table}

\begin{figure}[t]
\centering
\includegraphics[width=1\linewidth]{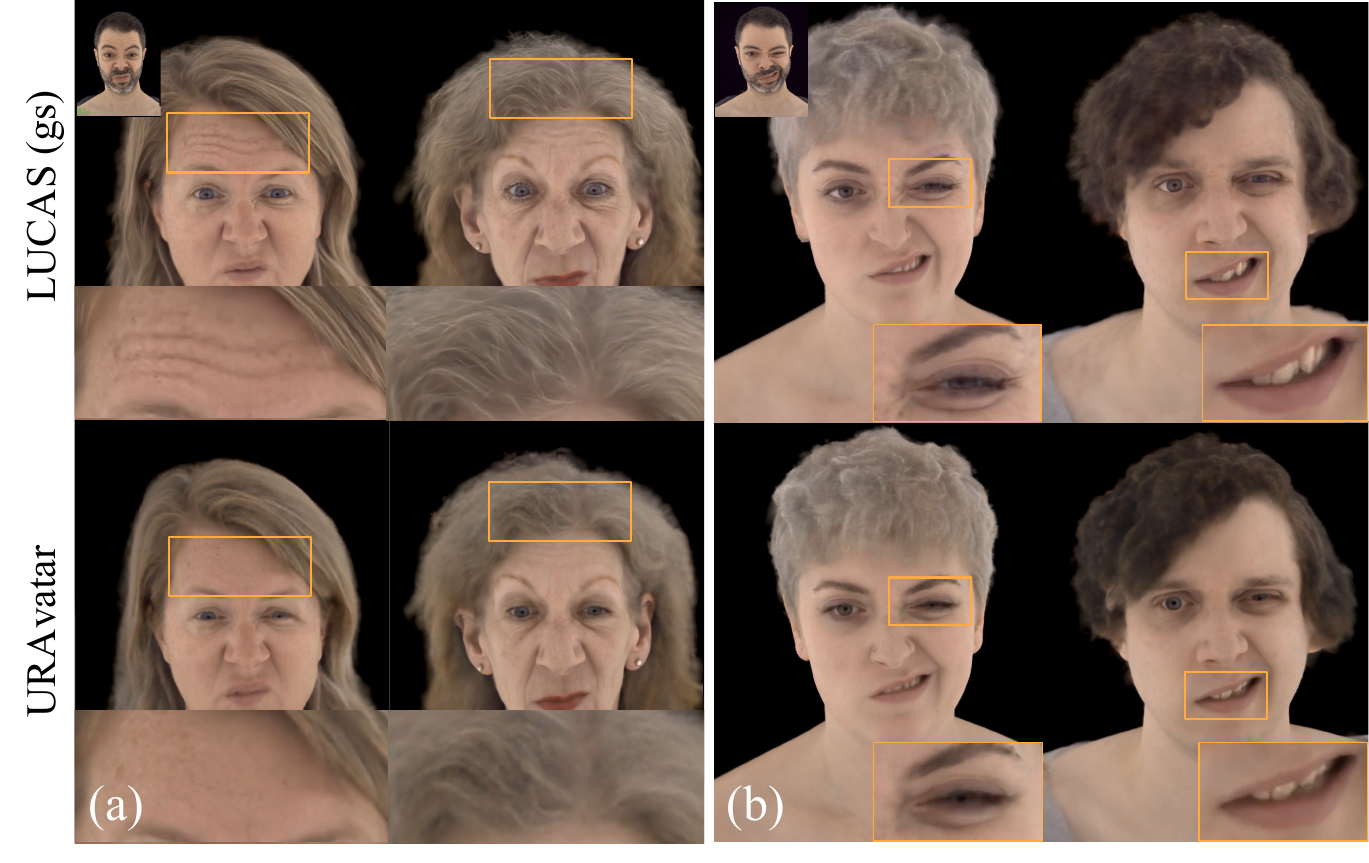}
\vspace{-15pt}
\caption{\textbf{Visualization of avatar driving.} (a) Expression retargeting from a source identity (top left) to multiple avatars demonstrates precise transfer of facial annd hair details. (b) Zero-shot driving on \textit{unseen subjects} shows accurate preservation of fine details around eyes and mouth regions.}
\label{fig:drive}
\end{figure}

\subsection{Evaluation of avatar driving}

% Our LUCAS avatars is drivable by various conditioning data. Expressions, poses, hairstyles.

% Fig. 1 right and Fig. 26 show some retargeting examples. Here, we choose one identity from our dataset (1st column in Fig. 26), pass the tracked mesh and texture into the expression encoder, to obtain the expres- sion code, and feed it into the decoder of each personalized avatar. These results show that the expression of the source identity is transferred to the different avatars, while details such as teeth and wrinkles are preserved.

\noindent \textbf{Driving avatars with diverse inputs.} Our universal model demonstrates versatile driving capabilities across different types of inputs as shown in Fig.\ref{cover_image}. More specifically, in Fig.~\ref{fig:drive}(a), expressions from a source identity (top left) are accurately transferred to multiple personalized avatars, preserving fine details in both wrinkles and hair. This precision stems from our universal layered prior model, where separate decoding of face and hair enables better reconstruction of intricate details.

% Additionally, we demonstrate hairstyle switching by changing the identity conditioning for the hair, showcasing the flexibility of our model to adapt different hairstyles seamlessly. 

\noindent \textbf{Testing on zero-shot driving.} To further evaluate generalization, we test our model on unseen subjects through zero-shot driving. Fig.~\ref{fig:drive}(b) demonstrates that our model successfully transfers novel expressions to untrained identities while maintaining precise facial features, particularly around the eyes and mouth regions.

\section{Conclusion}
We present LUCAS, the first universal compositional representation for 3D head avatars that disentangles face and hair components. This separation allows independent deformation, resolving issues like misplaced hair and misaligned dynamics. It also improves the anchor
geometry for precise and visually appealing Gaussian ren-
derings. Our Universal Layered Prior Model enables effective cross-identity generalization and avatar driving, even for unseen subjects.

\noindent \textbf{Limitation and future work.}
Our layered approach improves face and hair reconstruction but struggles with extreme hair deformations. Unseen poses during driving can degrade long hair deformation, especially in zero-shot scenarios. Future work will focus on relighting, training with a broader range of hairstyles for a more robust universal prior, and fine-tuning on real-world data to enhance applicability.

{
    \small
    \bibliographystyle{ieeenat_fullname}
    \bibliography{main}

\begin{thebibliography}{64}
\providecommand{\natexlab}[1]{#1}
\providecommand{\url}[1]{\texttt{#1}}
\expandafter\ifx\csname urlstyle\endcsname\relax
  \providecommand{\doi}[1]{doi: #1}\else
  \providecommand{\doi}{doi: \begingroup \urlstyle{rm}\Url}\fi

\bibitem[Abdi and Williams(2010)]{abdi2010principal}
Herv{\'e} Abdi and Lynne~J Williams.
\newblock Principal component analysis.
\newblock \emph{Wiley interdisciplinary reviews: computational statistics}, 2\penalty0 (4):\penalty0 433--459, 2010.

\bibitem[Bagautdinov et~al.(2018)Bagautdinov, Wu, Saragih, Fua, and Sheikh]{bagautdinov2018modeling}
Timur Bagautdinov, Chenglei Wu, Jason Saragih, Pascal Fua, and Yaser Sheikh.
\newblock Modeling facial geometry using compositional vaes.
\newblock In \emph{Proceedings of the IEEE Conference on Computer Vision and Pattern Recognition}, pages 3877--3886, 2018.

\bibitem[Blanz and Vetter(1999)]{Blanz1999AMM}
Volker Blanz and Thomas Vetter.
\newblock A morphable model for the synthesis of 3d faces.
\newblock \emph{Seminal Graphics Papers: Pushing the Boundaries, Volume 2}, 1999.

\bibitem[Blanz and Vetter(2023)]{blanz2023morphable}
Volker Blanz and Thomas Vetter.
\newblock A morphable model for the synthesis of 3d faces.
\newblock In \emph{Seminal Graphics Papers: Pushing the Boundaries, Volume 2}, pages 157--164. 2023.

\bibitem[Cao et~al.(2022)Cao, Simon, Kim, Schwartz, Zollhoefer, Saito, Lombardi, Wei, Belko, Yu, et~al.]{cao2022authentic}
Chen Cao, Tomas Simon, Jin~Kyu Kim, Gabriel Schwartz, Michael Zollhoefer, Shun-Suke Saito, Stephen Lombardi, Shih-En Wei, Danielle Belko, Shoou-I Yu, et~al.
\newblock Authentic volumetric avatars from a phone scan.
\newblock 2022.

\bibitem[Chai et~al.(2012)Chai, Wang, Weng, Yu, Guo, and Zhou]{chai2012single}
Menglei Chai, Lvdi Wang, Yanlin Weng, Yizhou Yu, Baining Guo, and Kun Zhou.
\newblock Single-view hair modeling for portrait manipulation.
\newblock \emph{ACM Transactions on Graphics (TOG)}, 31\penalty0 (4):\penalty0 1--8, 2012.

\bibitem[Chang et~al.(2022)Chang, Yan, Zhou, Liu, Sawalha, Ye, Zhangli, Kanski, Al’Aref, Axel, et~al.]{chang2022deeprecon}
Qi Chang, Zhennan Yan, Mu Zhou, Di Liu, Khalid Sawalha, Meng Ye, Qilong Zhangli, Mikael Kanski, Subhi Al’Aref, Leon Axel, et~al.
\newblock Deeprecon: Joint 2d cardiac segmentation and 3d volume reconstruction via a structure-specific generative method.
\newblock In \emph{International Conference on Medical Image Computing and Computer-Assisted Intervention}, pages 567--577. Springer, 2022.

\bibitem[Chu et~al.(2020)Chu, Ma, De~la Torre, Fidler, and Sheikh]{chu2020expressive}
Hang Chu, Shugao Ma, Fernando De~la Torre, Sanja Fidler, and Yaser Sheikh.
\newblock Expressive telepresence via modular codec avatars.
\newblock In \emph{Computer Vision--ECCV 2020: 16th European Conference, Glasgow, UK, August 23--28, 2020, Proceedings, Part XII 16}, pages 330--345. Springer, 2020.

\bibitem[Dao et~al.(2025)Dao, Doan, Liu, Le, and Metaxas]{dao2025improved}
Quan Dao, Khanh Doan, Di Liu, Trung Le, and Dimitris Metaxas.
\newblock Improved training technique for latent consistency models.
\newblock \emph{arXiv preprint arXiv:2502.01441}, 2025.

\bibitem[Feng et~al.(2023)Feng, Liu, Bolkart, Yang, Pollefeys, and Black]{feng2023learning}
Yao Feng, Weiyang Liu, Timo Bolkart, Jinlong Yang, Marc Pollefeys, and Michael~J Black.
\newblock Learning disentangled avatars with hybrid 3d representations.
\newblock \emph{arXiv preprint arXiv:2309.06441}, 2023.

\bibitem[Gao et~al.(2022)Gao, Zhou, Liu, Yan, Zhang, and Metaxas]{gao2022data}
Yunhe Gao, Mu Zhou, Di Liu, Zhennan Yan, Shaoting Zhang, and Dimitris~N Metaxas.
\newblock A data-scalable transformer for medical image segmentation: architecture, model efficiency, and benchmark.
\newblock \emph{arXiv preprint arXiv:2203.00131}, 2022.

\bibitem[Gao et~al.(2024)Gao, Li, Liu, Zhou, Zhang, and Metaxas]{gao2024training}
Yunhe Gao, Zhuowei Li, Di Liu, Mu Zhou, Shaoting Zhang, and Dimitris~N Metaxas.
\newblock Training like a medical resident: Context-prior learning toward universal medical image segmentation.
\newblock In \emph{Proceedings of the IEEE/CVF Conference on Computer Vision and Pattern Recognition}, pages 11194--11204, 2024.

\bibitem[Ghahramani and Hinton(1996)]{Ghahramani1996TheEA}
Zoubin Ghahramani and Geoffrey~E. Hinton.
\newblock The em algorithm for mixtures of factor analyzers.
\newblock 1996.

\bibitem[Ha et~al.(2016)Ha, Dai, and Le]{ha2016hypernetworks}
David Ha, Andrew Dai, and Quoc~V Le.
\newblock Hypernetworks.
\newblock \emph{arXiv preprint arXiv:1609.09106}, 2016.

\bibitem[Han et~al.(2024)Han, Wen, Chen, Zhang, Song, Ren, Gao, Stathopoulos, He, Chen, et~al.]{han2024proxedit}
Ligong Han, Song Wen, Qi Chen, Zhixing Zhang, Kunpeng Song, Mengwei Ren, Ruijiang Gao, Anastasis Stathopoulos, Xiaoxiao He, Yuxiao Chen, et~al.
\newblock Proxedit: Improving tuning-free real image editing with proximal guidance.
\newblock In \emph{Proceedings of the IEEE/CVF Winter Conference on Applications of Computer Vision}, pages 4291--4301, 2024.

\bibitem[He et~al.(2023)He, Tan, Liu, Si, Yao, Zhao, Liu, Zhangli, Chang, Li, et~al.]{he2023dealing}
Xiaoxiao He, Chaowei Tan, Bo Liu, Liping Si, Weiwu Yao, Liang Zhao, Di Liu, Qilong Zhangli, Qi Chang, Kang Li, et~al.
\newblock Dealing with heterogeneous 3d mr knee images: A federated few-shot learning method with dual knowledge distillation.
\newblock In \emph{2023 IEEE 20th International Symposium on Biomedical Imaging (ISBI)}, pages 1--5. IEEE, 2023.

\bibitem[He et~al.(2024)He, Han, Dao, Wen, Bai, Liu, Zhang, Min, Juefei-Xu, Tan, et~al.]{he2024dice}
Xiaoxiao He, Ligong Han, Quan Dao, Song Wen, Minhao Bai, Di Liu, Han Zhang, Martin~Renqiang Min, Felix Juefei-Xu, Chaowei Tan, et~al.
\newblock Dice: Discrete inversion enabling controllable editing for multinomial diffusion and masked generative models.
\newblock \emph{arXiv preprint arXiv:2410.08207}, 2024.

\bibitem[Kerbl et~al.(2023)Kerbl, Kopanas, Leimk{\"u}hler, and Drettakis]{kerbl20233d}
Bernhard Kerbl, Georgios Kopanas, Thomas Leimk{\"u}hler, and George Drettakis.
\newblock 3d gaussian splatting for real-time radiance field rendering.
\newblock \emph{ACM Trans. Graph.}, 42\penalty0 (4):\penalty0 139--1, 2023.

\bibitem[Kim et~al.(2024)Kim, Kim, Saito, and Joo]{kim2024gala}
Taeksoo Kim, Byungjun Kim, Shunsuke Saito, and Hanbyul Joo.
\newblock Gala: Generating animatable layered assets from a single scan.
\newblock In \emph{Proceedings of the IEEE/CVF Conference on Computer Vision and Pattern Recognition}, pages 1535--1545, 2024.

\bibitem[Kingma(2013)]{kingma2013auto}
Diederik~P Kingma.
\newblock Auto-encoding variational bayes.
\newblock \emph{arXiv preprint arXiv:1312.6114}, 2013.

\bibitem[Lewis et~al.(2014)Lewis, Anjyo, Rhee, Zhang, Pighin, and Deng]{lewis2014practice}
John~P Lewis, Ken Anjyo, Taehyun Rhee, Mengjie Zhang, Frederic~H Pighin, and Zhigang Deng.
\newblock Practice and theory of blendshape facial models.
\newblock \emph{Eurographics (State of the Art Reports)}, 1\penalty0 (8):\penalty0 2, 2014.

\bibitem[Li et~al.(2023)Li, Saito, Simon, Lombardi, Li, and Saragih]{li2023megane}
Junxuan Li, Shunsuke Saito, Tomas Simon, Stephen Lombardi, Hongdong Li, and Jason Saragih.
\newblock Megane: Morphable eyeglass and avatar network.
\newblock In \emph{Proceedings of the IEEE/CVF Conference on Computer Vision and Pattern Recognition}, pages 12769--12779, 2023.

\bibitem[Li et~al.(2024)Li, Cao, Schwartz, Khirodkar, Richardt, Simon, Sheikh, and Saito]{li2024uravatar}
Junxuan Li, Chen Cao, Gabriel Schwartz, Rawal Khirodkar, Christian Richardt, Tomas Simon, Yaser Sheikh, and Shunsuke Saito.
\newblock Uravatar: Universal relightable gaussian codec avatars.
\newblock \emph{arXiv preprint arXiv:2410.24223}, 2024.

\bibitem[Li et~al.(2017)Li, Bolkart, Black, Li, and Romero]{li2017learning}
Tianye Li, Timo Bolkart, Michael~J Black, Hao Li, and Javier Romero.
\newblock Learning a model of facial shape and expression from 4d scans.
\newblock \emph{ACM Trans. Graph.}, 36\penalty0 (6):\penalty0 194--1, 2017.

\bibitem[Lin et~al.(2024)Lin, Li, Su, Zheng, Zhang, and Liu]{lin2024layga}
Siyou Lin, Zhe Li, Zhaoqi Su, Zerong Zheng, Hongwen Zhang, and Yebin Liu.
\newblock Layga: Layered gaussian avatars for animatable clothing transfer.
\newblock In \emph{ACM SIGGRAPH 2024 Conference Papers}, pages 1--11, 2024.

\bibitem[Liu et~al.(2021{\natexlab{a}})Liu, Liu, Liu, Tao, Prince, and Carass]{liu2021label}
Di Liu, Jiang Liu, Yihao Liu, Ran Tao, Jerry~L Prince, and Aaron Carass.
\newblock Label super resolution for 3d magnetic resonance images using deformable u-net.
\newblock In \emph{Medical Imaging 2021: Image Processing}, pages 606--611. SPIE, 2021{\natexlab{a}}.

\bibitem[Liu et~al.(2021{\natexlab{b}})Liu, Yan, Chang, Axel, and Metaxas]{liu2021refined}
Di Liu, Zhennan Yan, Qi Chang, Leon Axel, and Dimitris~N Metaxas.
\newblock Refined deep layer aggregation for multi-disease, multi-view \& multi-center cardiac mr segmentation.
\newblock In \emph{International Workshop on Statistical Atlases and Computational Models of the Heart}, pages 315--322. Springer, 2021{\natexlab{b}}.

\bibitem[Liu et~al.(2022)Liu, Gao, Zhangli, Han, He, Xia, Wen, Chang, Yan, Zhou, et~al.]{liu2022transfusion}
Di Liu, Yunhe Gao, Qilong Zhangli, Ligong Han, Xiaoxiao He, Zhaoyang Xia, Song Wen, Qi Chang, Zhennan Yan, Mu Zhou, et~al.
\newblock Transfusion: multi-view divergent fusion for medical image segmentation with transformers.
\newblock In \emph{International conference on medical image computing and computer-assisted intervention}, pages 485--495. Springer, 2022.

\bibitem[Liu et~al.(2023{\natexlab{a}})Liu, Yu, Ye, Zhangli, Li, Zhang, and Metaxas]{liu2023deformer}
Di Liu, Xiang Yu, Meng Ye, Qilong Zhangli, Zhuowei Li, Zhixing Zhang, and Dimitris~N Metaxas.
\newblock Deformer: Integrating transformers with deformable models for 3d shape abstraction from a single image.
\newblock In \emph{Proceedings of the IEEE/CVF International Conference on Computer Vision}, pages 14236--14246, 2023{\natexlab{a}}.

\bibitem[Liu et~al.(2023{\natexlab{b}})Liu, Zhao, Zhangli, Gao, Liu, and Metaxas]{liu2023deep}
Di Liu, Long Zhao, Qilong Zhangli, Yunhe Gao, Ting Liu, and Dimitris~N Metaxas.
\newblock Deep deformable models: Learning 3d shape abstractions with part consistency.
\newblock \emph{arXiv preprint arXiv:2309.01035}, 2023{\natexlab{b}}.

\bibitem[Liu et~al.(2024{\natexlab{a}})Liu, Zhangli, Gao, and Metaxas]{liu2024lepard}
Di Liu, Qilong Zhangli, Yunhe Gao, and Dimitris Metaxas.
\newblock Lepard: Learning explicit part discovery for 3d articulated shape reconstruction.
\newblock \emph{Advances in Neural Information Processing Systems}, 36, 2024{\natexlab{a}}.

\bibitem[Liu et~al.(2024{\natexlab{b}})Liu, Zhuang, Metaxas, and Chandraker]{liu2024Instantaneous}
Di Liu, Bingbing Zhuang, Dimitris~N. Metaxas, and Manmohan Chandraker.
\newblock Instantaneous perception of moving objects in 3d.
\newblock In \emph{Proceedings of the IEEE/CVF conference on computer vision and pattern recognition}, 2024{\natexlab{b}}.

\bibitem[Lombardi et~al.(2018)Lombardi, Saragih, Simon, and Sheikh]{lombardi2018deep}
Stephen Lombardi, Jason Saragih, Tomas Simon, and Yaser Sheikh.
\newblock Deep appearance models for face rendering.
\newblock \emph{ACM Transactions on Graphics (ToG)}, 37\penalty0 (4):\penalty0 1--13, 2018.

\bibitem[Lombardi et~al.(2019)Lombardi, Simon, Saragih, Schwartz, Lehrmann, and Sheikh]{lombardi2019neural}
Stephen Lombardi, Tomas Simon, Jason Saragih, Gabriel Schwartz, Andreas Lehrmann, and Yaser Sheikh.
\newblock Neural volumes: Learning dynamic renderable volumes from images.
\newblock \emph{arXiv preprint arXiv:1906.07751}, 2019.

\bibitem[Lombardi et~al.(2021)Lombardi, Simon, Schwartz, Zollhoefer, Sheikh, and Saragih]{lombardi2021mixture}
Stephen Lombardi, Tomas Simon, Gabriel Schwartz, Michael Zollhoefer, Yaser Sheikh, and Jason Saragih.
\newblock Mixture of volumetric primitives for efficient neural rendering.
\newblock \emph{ACM Transactions on Graphics (ToG)}, 40\penalty0 (4):\penalty0 1--13, 2021.

\bibitem[Luo et~al.(2012)Luo, Li, Paris, Weise, Pauly, and Rusinkiewicz]{luo2012multi}
Linjie Luo, Hao Li, Sylvain Paris, Thibaut Weise, Mark Pauly, and Szymon Rusinkiewicz.
\newblock Multi-view hair capture using orientation fields.
\newblock In \emph{2012 IEEE Conference on Computer Vision and Pattern Recognition}, pages 1490--1497. IEEE, 2012.

\bibitem[Ma et~al.(2021)Ma, Simon, Saragih, Wang, Li, De~La~Torre, and Sheikh]{ma2021pixel}
Shugao Ma, Tomas Simon, Jason Saragih, Dawei Wang, Yuecheng Li, Fernando De~La~Torre, and Yaser Sheikh.
\newblock Pixel codec avatars.
\newblock In \emph{Proceedings of the IEEE/CVF Conference on Computer Vision and Pattern Recognition}, pages 64--73, 2021.

\bibitem[Mart{\'\i}n-Isla et~al.(2023)Mart{\'\i}n-Isla, Campello, Izquierdo, Kushibar, Sendra-Balcells, Gkontra, Sojoudi, Fulton, Arega, Punithakumar, et~al.]{martin2023deep}
Carlos Mart{\'\i}n-Isla, V{\'\i}ctor~M Campello, Cristian Izquierdo, Kaisar Kushibar, Carla Sendra-Balcells, Polyxeni Gkontra, Alireza Sojoudi, Mitchell~J Fulton, Tewodros~Weldebirhan Arega, Kumaradevan Punithakumar, et~al.
\newblock Deep learning segmentation of the right ventricle in cardiac mri: the m\&ms challenge.
\newblock \emph{IEEE Journal of Biomedical and Health Informatics}, 27\penalty0 (7):\penalty0 3302--3313, 2023.

\bibitem[Nicolet et~al.(2021)Nicolet, Jacobson, and Jakob]{Nicolet2021Large}
Baptiste Nicolet, Alec Jacobson, and Wenzel Jakob.
\newblock Large steps in inverse rendering of geometry.
\newblock \emph{ACM Transactions on Graphics (Proceedings of SIGGRAPH Asia)}, 40\penalty0 (6), 2021.

\bibitem[Paris et~al.(2008)Paris, Chang, Kozhushnyan, Jarosz, Matusik, Zwicker, and Durand]{paris2008hair}
Sylvain Paris, Will Chang, Oleg~I Kozhushnyan, Wojciech Jarosz, Wojciech Matusik, Matthias Zwicker, and Fr{\'e}do Durand.
\newblock Hair photobooth: geometric and photometric acquisition of real hairstyles.
\newblock \emph{ACM Trans. Graph.}, 27\penalty0 (3):\penalty0 30, 2008.

\bibitem[Pidhorskyi et~al.(2025)Pidhorskyi, Simon, Schwartz, Wen, Sheikh, and Saragih]{Pidhorskyi2024RasterizedEG}
Stanislav Pidhorskyi, Tomas Simon, Gabriel Schwartz, He Wen, Yaser Sheikh, and Jason Saragih.
\newblock Rasterized edge gradients: Handling discontinuities differentiably.
\newblock In \emph{Computer Vision -- ECCV 2024}, pages 335--352, Cham, 2025. Springer Nature Switzerland.

\bibitem[Raj et~al.(2021)Raj, Zollhofer, Simon, Saragih, Saito, Hays, and Lombardi]{raj2021pixel}
Amit Raj, Michael Zollhofer, Tomas Simon, Jason Saragih, Shunsuke Saito, James Hays, and Stephen Lombardi.
\newblock Pixel-aligned volumetric avatars.
\newblock In \emph{Proceedings of the IEEE/CVF Conference on Computer Vision and Pattern Recognition}, pages 11733--11742, 2021.

\bibitem[Ranjan et~al.(2018)Ranjan, Bolkart, Sanyal, and Black]{ranjan2018generating}
Anurag Ranjan, Timo Bolkart, Soubhik Sanyal, and Michael~J Black.
\newblock Generating 3d faces using convolutional mesh autoencoders.
\newblock In \emph{Proceedings of the European conference on computer vision (ECCV)}, pages 704--720, 2018.

\bibitem[Richard et~al.(2021)Richard, Lea, Ma, Gall, De~la Torre, and Sheikh]{richard2021audio}
Alexander Richard, Colin Lea, Shugao Ma, Jurgen Gall, Fernando De~la Torre, and Yaser Sheikh.
\newblock Audio-and gaze-driven facial animation of codec avatars.
\newblock In \emph{Proceedings of the IEEE/CVF winter conference on applications of computer vision}, pages 41--50, 2021.

\bibitem[Saito et~al.(2024)Saito, Schwartz, Simon, Li, and Nam]{saito2024relightable}
Shunsuke Saito, Gabriel Schwartz, Tomas Simon, Junxuan Li, and Giljoo Nam.
\newblock Relightable gaussian codec avatars.
\newblock In \emph{Proceedings of the IEEE/CVF Conference on Computer Vision and Pattern Recognition}, pages 130--141, 2024.

\bibitem[Schwartz et~al.(2020)Schwartz, Wei, Wang, Lombardi, Simon, Saragih, and Sheikh]{schwartz2020eyes}
Gabriel Schwartz, Shih-En Wei, Te-Li Wang, Stephen Lombardi, Tomas Simon, Jason Saragih, and Yaser Sheikh.
\newblock The eyes have it: An integrated eye and face model for photorealistic facial animation.
\newblock \emph{ACM Transactions on Graphics (TOG)}, 39\penalty0 (4):\penalty0 91--1, 2020.

\bibitem[Sevastopolsky et~al.(2025)Sevastopolsky, Grassal, Giebenhain, Athar, Verdoliva, and Neissner]{sevastopolsky2025headcraft}
Artem Sevastopolsky, Philip-William Grassal, Simon Giebenhain, Shah{R}ukh Athar, Luisa Verdoliva, and Matthias Neissner.
\newblock Headcraft: Modeling high-detail shape variations for animated 3dmms.
\newblock 2025.

\bibitem[Shamai et~al.(2019)Shamai, Slossberg, and Kimmel]{shamai2019synthesizing}
Gil Shamai, Ron Slossberg, and Ron Kimmel.
\newblock Synthesizing facial photometries and corresponding geometries using generative adversarial networks.
\newblock \emph{ACM Transactions on Multimedia Computing, Communications, and Applications (TOMM)}, 15\penalty0 (3s):\penalty0 1--24, 2019.

\bibitem[Sorkine-Hornung and Alexa(2007)]{SorkineHornung2007AsrigidaspossibleSM}
Olga Sorkine-Hornung and Marc Alexa.
\newblock As-rigid-as-possible surface modeling.
\newblock In \emph{Eurographics Symposium on Geometry Processing}, 2007.

\bibitem[Torresani et~al.(2003)Torresani, Hertzmann, and Bregler]{NIPS2003_8db92642}
Lorenzo Torresani, Aaron Hertzmann, and Christoph Bregler.
\newblock Learning non-rigid 3d shape from 2d motion.
\newblock In \emph{Advances in Neural Information Processing Systems}. MIT Press, 2003.

\bibitem[Wang et~al.(2019)Wang, Sun, Cheng, Jiang, Deng, Zhao, Liu, Mu, Tan, Wang, Liu, and Xiao]{Wang2019DeepHR}
Jingdong Wang, Ke Sun, Tianheng Cheng, Borui Jiang, Chaorui Deng, Yang Zhao, Dong Liu, Yadong Mu, Mingkui Tan, Xinggang Wang, Wenyu Liu, and Bin Xiao.
\newblock Deep high-resolution representation learning for visual recognition.
\newblock \emph{IEEE Transactions on Pattern Analysis and Machine Intelligence}, 43:\penalty0 3349--3364, 2019.

\bibitem[Wang et~al.(2004)Wang, Bovik, Sheikh, and Simoncelli]{wang2004image}
Zhou Wang, Alan~C Bovik, Hamid~R Sheikh, and Eero~P Simoncelli.
\newblock Image quality assessment: from error visibility to structural similarity.
\newblock \emph{IEEE transactions on image processing}, 13\penalty0 (4):\penalty0 600--612, 2004.

\bibitem[Wang et~al.(2022)Wang, Nam, Stuyck, Lombardi, Zollh{\"o}fer, Hodgins, and Lassner]{wang2022hvh}
Ziyan Wang, Giljoo Nam, Tuur Stuyck, Stephen Lombardi, Michael Zollh{\"o}fer, Jessica Hodgins, and Christoph Lassner.
\newblock Hvh: Learning a hybrid neural volumetric representation for dynamic hair performance capture.
\newblock In \emph{Proceedings of the IEEE/CVF Conference on Computer Vision and Pattern Recognition}, pages 6143--6154, 2022.

\bibitem[Wang et~al.(2023)Wang, Nam, Stuyck, Lombardi, Cao, Saragih, Zollh{\"o}fer, Hodgins, and Lassner]{wang2023neuwigs}
Ziyan Wang, Giljoo Nam, Tuur Stuyck, Stephen Lombardi, Chen Cao, Jason Saragih, Michael Zollh{\"o}fer, Jessica Hodgins, and Christoph Lassner.
\newblock Neuwigs: A neural dynamic model for volumetric hair capture and animation.
\newblock In \emph{Proceedings of the IEEE/CVF Conference on Computer Vision and Pattern Recognition}, pages 8641--8651, 2023.

\bibitem[Wei et~al.(2019)Wei, Saragih, Simon, Harley, Lombardi, Perdoch, Hypes, Wang, Badino, and Sheikh]{wei2019vr}
Shih-En Wei, Jason Saragih, Tomas Simon, Adam~W Harley, Stephen Lombardi, Michal Perdoch, Alexander Hypes, Dawei Wang, Hernan Badino, and Yaser Sheikh.
\newblock Vr facial animation via multiview image translation.
\newblock \emph{ACM Transactions on Graphics (ToG)}, 38\penalty0 (4):\penalty0 1--16, 2019.

\bibitem[Wu et~al.(2024)Wu, Yang, Kuang, Feng, Han, Shen, Fu, Zhou, and Zheng]{wu2024monohair}
Keyu Wu, Lingchen Yang, Zhiyi Kuang, Yao Feng, Xutao Han, Yuefan Shen, Hongbo Fu, Kun Zhou, and Youyi Zheng.
\newblock Monohair: High-fidelity hair modeling from a monocular video.
\newblock In \emph{Proceedings of the IEEE/CVF Conference on Computer Vision and Pattern Recognition}, pages 24164--24173, 2024.

\bibitem[Xu et~al.(2023)Xu, Zhang, Wang, Zhao, Huang, Qi, and Liu]{xu2023latentavatar}
Yuelang Xu, Hongwen Zhang, Lizhen Wang, Xiaochen Zhao, Han Huang, Guojun Qi, and Yebin Liu.
\newblock Latentavatar: Learning latent expression code for expressive neural head avatar.
\newblock In \emph{ACM SIGGRAPH 2023 Conference Proceedings}, pages 1--10, 2023.

\bibitem[Zhang et~al.(2024)Zhang, Feng, Kulits, Wen, Thies, and Black]{zhang2024teca}
Hao Zhang, Yao Feng, Peter Kulits, Yandong Wen, Justus Thies, and Michael~J Black.
\newblock Teca: Text-guided generation and editing of compositional 3d avatars.
\newblock In \emph{2024 International Conference on 3D Vision (3DV)}, pages 1520--1530. IEEE, 2024.

\bibitem[Zhang et~al.(2018)Zhang, Isola, Efros, Shechtman, and Wang]{zhang2018unreasonable}
Richard Zhang, Phillip Isola, Alexei~A Efros, Eli Shechtman, and Oliver Wang.
\newblock The unreasonable effectiveness of deep features as a perceptual metric.
\newblock In \emph{Proceedings of the IEEE conference on computer vision and pattern recognition}, pages 586--595, 2018.

\bibitem[Zhangli et~al.(2022)Zhangli, Yi, Liu, He, Xia, Chang, Han, Gao, Wen, Tang, et~al.]{zhangli2022region}
Qilong Zhangli, Jingru Yi, Di Liu, Xiaoxiao He, Zhaoyang Xia, Qi Chang, Ligong Han, Yunhe Gao, Song Wen, Haiming Tang, et~al.
\newblock Region proposal rectification towards robust instance segmentation of biological images.
\newblock In \emph{International Conference on Medical Image Computing and Computer-Assisted Intervention}, pages 129--139. Springer, 2022.

\bibitem[Zhangli et~al.(2024)Zhangli, Jiang, Liu, Yu, Dai, Ramchandani, Pang, Metaxas, and Krishnan]{zhangli2024layout}
Qilong Zhangli, Jindong Jiang, Di Liu, Licheng Yu, Xiaoliang Dai, Ankit Ramchandani, Guan Pang, Dimitris~N Metaxas, and Praveen Krishnan.
\newblock Layout-agnostic scene text image synthesis with diffusion models.
\newblock In \emph{2024 IEEE/CVF Conference on Computer Vision and Pattern Recognition (CVPR)}, pages 7496--7506. IEEE Computer Society, 2024.

\bibitem[Zheng et~al.(2023)Zheng, Jin, Li, Huang, Ma, Cui, and Han]{zheng2023hairstep}
Yujian Zheng, Zirong Jin, Moran Li, Haibin Huang, Chongyang Ma, Shuguang Cui, and Xiaoguang Han.
\newblock Hairstep: Transfer synthetic to real using strand and depth maps for single-view 3d hair modeling.
\newblock In \emph{Proceedings of the IEEE/CVF Conference on Computer Vision and Pattern Recognition}, pages 12726--12735, 2023.

\bibitem[Zhou et~al.(2018)Zhou, Hu, Xing, Chen, Kung, Tong, and Li]{zhou2018hairnet}
Yi Zhou, Liwen Hu, Jun Xing, Weikai Chen, Han-Wei Kung, Xin Tong, and Hao Li.
\newblock Hairnet: Single-view hair reconstruction using convolutional neural networks.
\newblock In \emph{Proceedings of the European Conference on Computer Vision (ECCV)}, pages 235--251, 2018.

\bibitem[Zhou et~al.(2019)Zhou, Deng, Kotsia, and Zafeiriou]{zhou2019dense}
Yuxiang Zhou, Jiankang Deng, Irene Kotsia, and Stefanos Zafeiriou.
\newblock Dense 3d face decoding over 2500fps: Joint texture \& shape convolutional mesh decoders.
\newblock In \emph{Proceedings of the IEEE/CVF conference on computer vision and pattern recognition}, pages 1097--1106, 2019.

\end{thebibliography}
}
% \newpage

\end{document}

\maketitle

\section{Network Architecture}
In this section, we provide more details of our network architecture and hyperparameters for our id-conditioned hypernetwork, appearance deocder, geometry decoder, Gaussian hypernetwork and Gaussian decoder, respectively.

\noindent \textbf{Identity-conditioned hypernetwork.}
We adopt a U-Net~\cite{ronneberger2015u} architecture as our identity-conditioned hypernetwork that takes as input the neutral geometry and texture of a subject to predict subject-specific decoder parameters. The network consists of two parallel downsampling branches that process geometry and texture features separately, followed by a joint upsampling branch. The input geometry and texture are first normalized by subtracting their respective means and dividing by their standard deviations. The geometry branch processes the normalized geometry (scaled by 0.2) while the texture branch processes the normalized texture (scaled by 0.4) through a series of downsampling blocks. 
The network's channel dimensions progressively increase through the layers with sizes of (3, 32, 64, 128, 256, 256, 512, 512, 512, 512, 256). Features from both branches are concatenated at each scale and processed through the upsampling branch. The network generates three types of outputs through Transfer modules (Fig.~\ref{fig:supp_block}(a)): untied bias parameters ($\Theta_{\text{id}}$), per-identity geometry displacement ($d$), and per-identity positional encoding ($f$). Each Transfer module consists of two weight-normalized convolution (Conv2DWN) layers with learnable biases, mapping the concatenated features through a hidden dimension of 512 channels, followed by LeakyReLU activation ($\alpha=0.2$), before outputting the final parameters.
For both face and hair branches, six texture bias parameters are generated for the appearance decoder, corresponding to different resolution levels. The geometry decoder receives five geometry displacement parameters for its Block modules, with an additional displacement map ($d$) for the output layer. The pixel decoder receives a positional encoding ($f$) of dimension [4, 1024, 1024] generated by processing the concatenation of the final upsampled features and the input texture through a Transfer module.
The architecture leverages weight normalization throughout its convolution and linear layers for stable training, with careful initialization using Glorot initialization scaled by 0.2 for most layers and 1.0 for the final convolution layers in the Transfer modules.

\begin{figure}[t]
\centering
\includegraphics[width=1\linewidth]{figs/supp_block.pdf}
\caption{\textbf{Architecture of (a) Transfer module and (b) Block module.} (a) Transfer Block processes features through two Conv2dWN layers with intermediate LeakyReLU activation and scaled learnable bias. (b) Block module combines Conv2dWN, LeakyReLU activation, and pixel shuffle operations for feature transformation and upsampling.}
\label{fig:supp_block}
\end{figure}

\noindent \textbf{Appearance decoder.}
The face appearance decoder $\mathcal{D}_e^\text{face}$ processes expression encoding $z$, view-dependent conditions $\omega$ and neck pose $\eta$ to generate detailed textures. Specifically, $z \in \mathbb{R}^{16}$ is first processed by a Block module Fig.~\ref{fig:supp_block}(b)) to obtain features in $\mathbb{R}^{128}$, while the view direction encoding $\omega \in \mathbb{R}^3$ and neck pose $\eta \in \mathbb{R}^6$ are transformed through a linear layer and LeakyReLU activation to $\mathbb{R}^{16}$. These combined features are then processed through a cascade of Block modules that progressively upsample the spatial resolution from $8\times8$ to $256\times256$ while reducing the channel dimensions ($160, 64, 32, 16, 12, 8, 4$). The view-dependent conditioning enables the network to capture view-dependent effects like specular highlights and shading variations. For hair appearance decoder $\mathcal{D}_e^\text{hair}$, we use a similar network architecture with additional head pose $h$ encoded to $\mathbb{R}^{16}$ through linear layers and LeakyReLU activations, resulting in a $176$-dimensional feature vector after concatenation.

\noindent \textbf{Geometry decoder.}
The face geometry decoder $\mathcal{D}_g^\text{face}$ combines expression encoding $z$ with neck pose parameters ${\eta}$ to capture expression-dependent geometry deformations and neck movements. The input $z \in \mathbb{R}^{16}$ is first processed by a Block module, expanding the features to $\mathbb{R}^{128}$. Meanwhile, the neck pose ${\eta}$ is encoded through a linear layer and LeakyReLU to $\mathbb{R}^{16}$. These concatenated features are processed through the same progressive upsampling architecture as the appearance decoder. Notably, the geometry decoder includes a mask output that helps handle occlusions or invalid regions that may occur due to extreme neck poses or expressions. For hair geometry decoder $\mathcal{D}_g^\text{hair}$, the network additionally conditions on head pose $h$, which is encoded similarly to $\eta$, resulting in a $160$-dimensional feature vector after concatenation.

\begin{figure}[!htbp]
\centering
\includegraphics[width=1\linewidth]{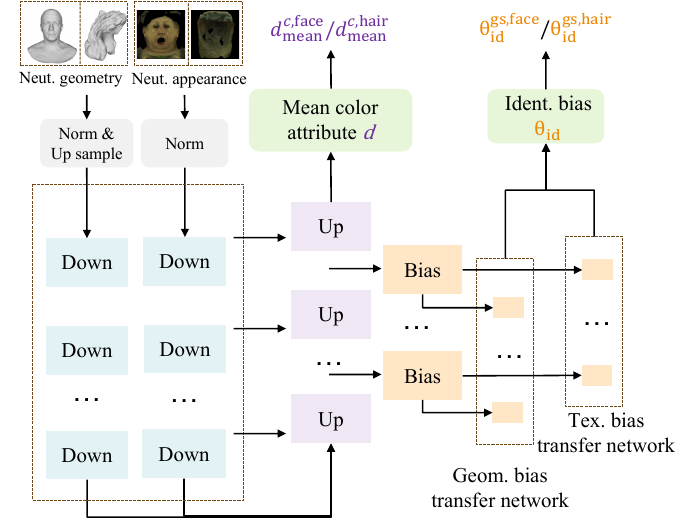}
\caption{\textbf{Overview of Gaussian hypernetwork.} The network processes normalized geometry $\textbf{G}_{\text{neu}}$ and texture $\textbf{T}_{\text{neu}}$ through parallel branches to predict identity-specific bias parameters $\Theta_\text{id}^{\text{gs}}$ and mean color attributes $d^c_\text{mean}$ for Gaussian rendering. The same architecture is used for both face and hair branches.}
\label{fig:supp_framework_gs_hyper}
\end{figure}

\noindent \textbf{Gaussian-splatting hypernetwork.}
As shown in Fig.~\ref{fig:supp_framework_gs_hyper}, our Gaussian hypernetwork $\mathcal{E}_\text{id}^\text{gs}$ shares the same U-Net backbone as the PiCA hypernetwork, consisting of two parallel downsampling branches that process geometry $\textbf{G}_{\text{neu}}$ and texture $\textbf{T}_{\text{neu}}$ features separately. Through the dedicated downsampling branches, the input geometry and texture are first normalized ($0.2\times$ and $0.4\times$ respectively) and processed into multi-scale feature maps. The resulting features are concatenated at each scale and processed through an upsampling network. Through a sequence of transfer modules, the network predicts the identity-specific bias map $\Theta_\text{id}^{\text{gs}}$ and extracts mean color attributes $d^c_\text{mean}$ from the neutral appearance data, as formulated in Eq.~{\color{red}7} of the main paper. These bias parameters control the position, rotation, scale, opacity and color attribute of the splatted Gaussians anchored at mesh vertices $\{\hat{t}_k\}_{k=1}^M$, allowing for person-specific rendering characteristics. The same architecture is employed for both face and hair branches, enabling joint optimization of all rendering components through a unified hypernetwork backbone.

\noindent \textbf{Gaussian decoder.} 
The Gaussian decoder $\mathcal{D}^{\text{gs}}$ consists of a progressive upsampling network that transforms input features to Gaussian attributes. Taking as input a 128-dimensional expression encoding concatenated with a 16-dimensional neck pose encoding $\eta$, the network first processes them through two MLP layers to obtain $8\times8$ feature maps. These features are then gradually upsampled through a series of transposed convolution layers with LeakyReLU activations ($\alpha=0.2$), expanding the spatial resolution from $8\times8$ to $1024\times1024$ while progressively reducing the channel dimensions ($256, 128, 64, 32, 16, 59$). The final layer outputs a 59-channel feature map, where the first 49 channels encode the spherical harmonics coefficients for appearance, and the remaining 10 channels encode the Gaussian attributes: position delta $\delta t_k$, rotation quaternion $q_k$, and scale $s_k$. The quaternions are normalized and scales are constrained through a softplus activation multiplied by 0.5. An additional sigmoid activation is applied to the first spherical harmonic coefficient to obtain the opacity $o_k$. The face and hair branches share the same architecture but operate independently to handle their respective geometry and appearance characteristics.

\noindent \textbf{Gaussian Rendering.} 
After obtaining the Gaussian attributes from both face and hair decoders, we concatenate their features for joint rendering. Let $N = N_f + N_h$ denote the total number of Gaussians, where $N_f$ and $N_h$ represent face and hair Gaussians respectively. We combine their position deltas $\delta t \in \mathbb{R}^{N\times3}$, rotation quaternions $q \in \mathbb{R}^{N\times4}$, scales $s \in \mathbb{R}^{N\times3}$, opacities $o \in \mathbb{R}^{N\times1}$, and spherical harmonics coefficients $d^c \in \mathbb{R}^{N\times48}$. 
For RGB rendering, the color attribute $c_k \in \mathbb{R}^{48}$ for the $k$-th Gaussian is computed as:
\begin{equation}
c_k = 
\begin{bmatrix}
    d_k^{c,\text{base}} + d^c_\text{mean} \\
     \beta \cdot d_k^{c,\text{ho}}
\end{bmatrix}
\end{equation}
where $d_k^{c,\text{base}} \in \mathbb{R}^3 $ is the base color component (first three coefficients), $d_{\text{mean}}^c \in \mathbb{R}^3 $ is the mean color vector, $d_k^{c,\text{ho}} \in \mathbb{R}^{45}$ represents the higher-order coefficients (remaining 45 coefficients). $\beta$ is a scaling factor for the higher-order terms where we set as 0.05 in our experiment.
% For segmentation mask rendering, we assign face and hair Gaussians with different ID values (1 for face, 2 for hair) to obtain per-pixel part labels. The rendering is performed using a GPU-accelerated EGL-based OpenGL renderer, which rasterizes each Gaussian splat according to its 3D position, scale, and rotation under the given camera parameters, and composites them in a front-to-back order to generate the final RGB image or segmentation mask.

\section{Training details}
% We first train personalized model for all identities. Then we calculate the average geometry and texture for each identity and take them as static assets for the universal prior model training.

% \noindent \textbf{Training strategy.} 
Our model follows a two-phase training process. In the first phase, we train only the PiCA branch using reconstruction loss $\mathcal{L}_{\text{pica}}$ and dehairing loss $\mathcal{L}_{\text{dehair}}$. Subsequently, we freeze these weights and train the Gaussian branch with loss $\mathcal{L}_{\text{gs}}$.
The loss weights are configured as follows:
\begin{itemize}
    \item Gaussian loss $\mathcal{L}_{\text{gs}}$: $\lambda_{\text{render}}$ = 10.0, $\lambda_{\text{scale}}$ = 0.2, $\lambda_{\Delta}$ = 0.01;
    \item PiCA loss $\mathcal{L}_{\text{pica}}$: $\lambda_I$ = 4.0, $\lambda_D$ = 10.0, $\lambda_{N}$ = 1.0, $\lambda_{M}$ = 0.1, $\lambda_{S}$ = 4.0, $\lambda_{KL}$ = 0.001, $\lambda_{seg}$ = 2.0;
    \item Dehairing loss $\mathcal{L}_{\text{dehair}}$: initially $\lambda_{\text{dehair}}$ = 20.0, decaying to 0 between 70k and 80k iterations;
    \item $\lambda_{\text{pica}}$ = $\lambda_{\text{gs}}$ = $\lambda_{\text{dehair}}$ = 1.0.
\end{itemize}
We adopt L1 loss for both $\mathcal{L}_{I}$ and $\mathcal{L}_{\text{render}}$ due to its effectiveness in preserving fine hair details. Hair segmentation regularization $\mathcal{L}_{\text{seg}}$ is restricted to the first phase to avoid Gaussian blur artifacts between hair strands. The high initial weight of dehairing loss $\mathcal{L}_{\text{dehair}}$ accelerates the convergence of bald geometry, ensuring accurate dehaired results without interference from the hair mesh. The model is trained using Adam optimizer with a learning rate of 0.001 for 300k (PiCA branch) + 300k  (GS branch) iterations on 4 NVIDIA A100 GPUs.
\begin{figure}[t]
\centering
\includegraphics[width=1\linewidth]{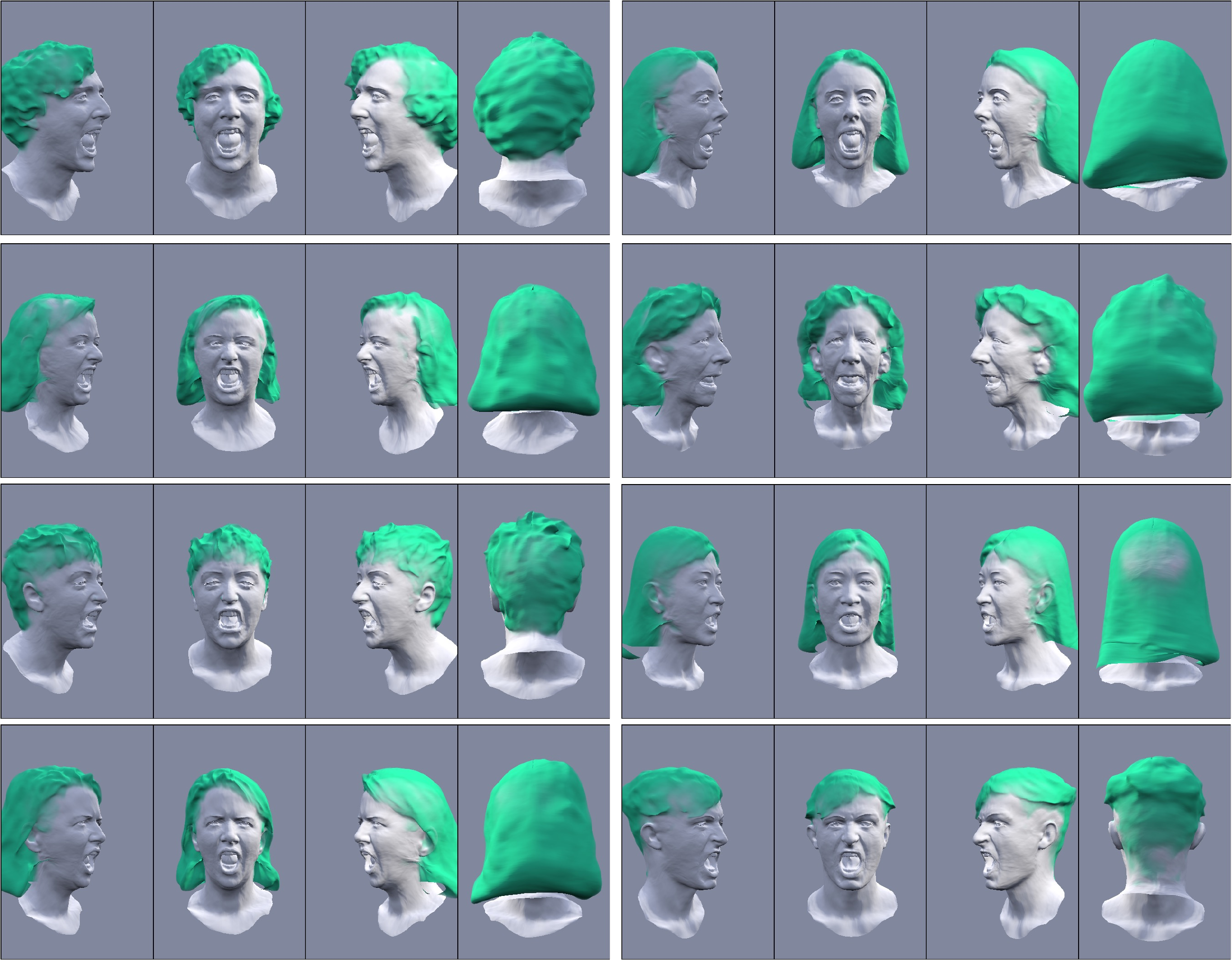}
\caption{\textbf{More visualization of avatar dehairing.} Our method successfully removes hair while preserving the underlying head geometry across subjects with diverse hairstyles. }
\label{fig:supp_dehair}
\end{figure}

\section{Discussion}
\noindent \textbf{Avatar dehairing.}
We provide more visualization of the dehairing results in Fig.~\ref{fig:supp_dehair}. The examples demonstrate that our method can effectively remove diverse hairstyles while maintaining accurate head shape and facial features.

\noindent \textbf{Impact of training strategy.} We investigate two different training strategies for our Gaussian model: two-stage training and joint training. In the two-stage approach, we first train the mesh model with $\mathcal{L}_\text{pica}$ and $\mathcal{L}_\text{dehair}$ for 300k iterations, then freeze the mesh branch and train the Gaussian branch with $\mathcal{L}_\text{pica}$ for another 300k iterations. In the joint training approach, we train both branches simultaneously with all losses enabled. As shown in Table~\ref{tab:training_strategy}, both strategies achieve comparable performance across all metrics, with differences being negligible (PSNR: ±0.003, SSIM: ±0.0005, LPIPS: ±0.0007). However, we observe that joint training exhibits significant instability during the initial 10k iterations, often leading to training explosions that substantially delay the convergence process. Given these findings, we adopt the two-stage training strategy in our final model for its superior training stability while maintaining equivalent performance.
\begin{table}[t]
\centering
\caption{\textbf{Ablation study} on training strategy. While both strategies achieve similar final performance, two-stage training offers better stability during the optimization process.
}

\renewcommand\tabcolsep{12pt}
\resizebox{1\linewidth}{!}{
\begin{tabular}{lccc}
\toprule
Strategy & PSNR $\uparrow$ & SSIM $\uparrow$ & LPIPS $\downarrow$ \\ 
\midrule
Two-stage training & \textbf{34.5607} & 0.9196 & \textbf{0.2387} \\
Joint training &{34.5579} & \textbf{0.9201} & {0.2394} \\
\bottomrule
\end{tabular}}
\label{tab:training_strategy}
\end{table}
\begin{figure}[t]
\centering
\includegraphics[width=1\linewidth]{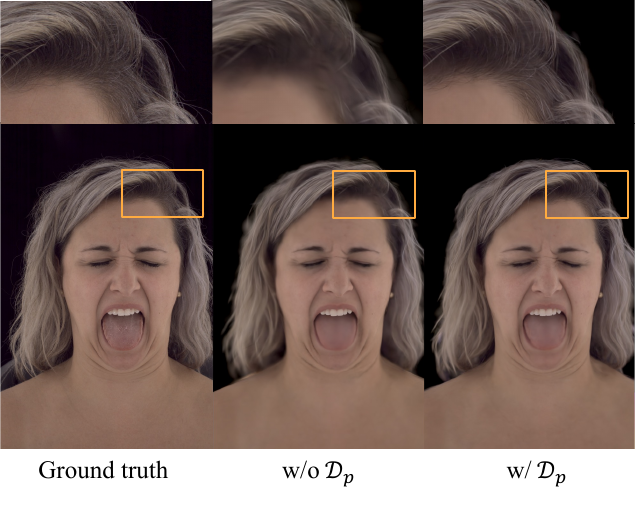}

\caption{\textbf{Impact of pixel decoder $\mathcal{D}_p$.} The pixel decoder helps achieve better facial details and overall fidelity.}
\label{fig:supp_dp}
\end{figure}

\noindent \textbf{Impact of pixel decoder.} To understand the role of the pixel decoder in our Gaussian model, we conduct experiments by removing $\mathcal{D}_p$ and supervising the reconstruction solely through Gaussian rendering loss. Note that this ablation is conducted with joint training of the mesh and Gaussian branches. As shown in Fig.~\ref{fig:supp_dp}, models with $\mathcal{D}_p$ achieve noticeably better visual fidelity compared to those without. The quantitative results in Table.~\ref{tab:ab_p} further support this observation, with our full model outperforming the variant without $\mathcal{D}_p$ across most metrics, particularly on unseen subjects. We attribute this improvement to the pixel decoder's ability to enhance the underlying avatar geometry, which in turn provides better spatial anchoring for Gaussian rendering. Notably, the performance gap widens on unseen subjects, suggesting that $\mathcal{D}_p$ contributes to better generalization of our model. Furthermore, even without the pixel decoder, our model significantly outperforms URAvatar~\cite{li2024uravatar}, demonstrating the inherent advantage of our layered representation design.

\begin{table}[t]
\centering
\caption{\textbf{Ablation study} on pixel decoder $\mathcal{D}_p$. We evaluate our universal model on both training and unseen subjects. The top two techniques are highlighted in \textcolor{red!50}{red} and \textcolor{yellow!100}{yellow}, respectively.}

\renewcommand\tabcolsep{2pt}
\resizebox{1\linewidth}{!}{
\begin{tabular}{lcccccc}
\toprule
& \multicolumn{3}{c}{Training subjects} & \multicolumn{3}{c}{Unseen subjects} \\
\cmidrule(lr){2-4} \cmidrule(lr){5-7}
\multirow{1}{*}{Method}  & PSNR $\uparrow$ & SSIM $\uparrow$ & LPIPS $\downarrow$ & PSNR $\uparrow$ & SSIM $\uparrow$ & LPIPS $\downarrow$ \\ 
\midrule
w/o $\mathcal{D}_p$ & \cellcolor{yellow!25}34.2113 & \cellcolor{yellow!25}0.9164 & \cellcolor{red!25}0.2390 & \cellcolor{yellow!25}32.1328 & \cellcolor{yellow!25}0.8993 & \cellcolor{yellow!25}0.2607 \\

URAvatar & 33.1209 & 0.9021 & 0.2493 & 31.4462 & 0.8922 & 0.2625\\

Full model & \cellcolor{red!25}{34.5579} & \cellcolor{red!25}{0.9201} & \cellcolor{yellow!25}{0.2394}  & \cellcolor{red!25}{32.5847} & \cellcolor{red!25}{0.9087} & \cellcolor{red!25}{0.2496} \\
\bottomrule
\end{tabular}}

\label{tab:ab_p}
\end{table}

\begin{figure}[t]
\centering
\includegraphics[width=1\linewidth]{figs/supp_gs_delta.pdf}

\caption{\textbf{Impact of Gaussian position prior regularization.} The position delta loss $\mathcal{L}_\Delta$ effectively constrains the Gaussians to stay within their respective regions, leading to clean boundaries between hair and bald head areas and faithful reproduction of the ground truth appearance.}
\label{fig:supp_gs_delta}
\end{figure}

% \begin{table}[t]
% \centering
% \caption{Ablation study on the number of training subjects. The top three techniques are highlighted in \textcolor{red!50}{red}, \textcolor{orange!50}{orange}, and \textcolor{yellow!100}{yellow}, respectively. }
% 
% \renewcommand\tabcolsep{17pt}
% \resizebox{1\linewidth}{!}{
% \begin{tabular}{lccc}
% \toprule
% Method & PSNR $\uparrow$ & SSIM $\uparrow$ & LPIPS $\downarrow$ \\ 
% \midrule
% 4 & 30.5284 & 0.8749 & 0.2690\\
% 8 & 31.5512 & 0.8825 & 0.2608   \\
% 16 & 32.0852 & 0.9022 & 0.2557\\
% 32 & 33.5858 & 0.9089 & 0.2492 \\
% 48 & 33.9327 & 0.9136 & 0.2457 \\
% 64 & 34.2785 & 0.9175 & 0.2424\\
% 76 & 34.5579  & 0.9201 & 0.2394\\

% \bottomrule
% \end{tabular}}
% 
% \label{tab:subject_number}
% \end{table}

% \begin{table}[t]
% \centering
% \caption{Ablation study on the number of training subjects. The top three techniques are highlighted in \textcolor{red!50}{red}, \textcolor{orange!50}{orange}, and \textcolor{yellow!100}{yellow}, respectively. }
% 
% \renewcommand\tabcolsep{17pt}
% \resizebox{1\linewidth}{!}{
% \begin{tabular}{lccc}
% \toprule
% Method & PSNR $\uparrow$ & SSIM $\uparrow$ & LPIPS $\downarrow$ \\ 
% \midrule
% 4 \\
% 8 & 32.0512 & 0.8895 & 0.2678   \\
% 16 \\
% 32 \\
% 48 \\
% 64 \\
% 76 & 33.0254 & 0.9073  & 0.2537 \\

% \bottomrule
% \end{tabular}}
% 
% \label{tab:subject_number}
% \end{table}

\noindent \textbf{Impact of Gaussian position prior regularization.} In Fig.~\ref{fig:supp_gs_delta} we show the Gaussian shapes for the hair and bald head regions. With the position delta loss $\mathcal{L}_\Delta$, our model effectively maintains the Gaussian primitives in their designated regions - hair Gaussians properly represent the hair volume while face Gaussians accurately cover the bald head area. As shown in the visualization, the Gaussian distributions align well with the ground truth appearance, leading to high-quality rendered results. The clear separation between hair and bald head Gaussians demonstrates the effectiveness of our position regularization.

\begin{table}[t]
\centering
\caption{\textbf{Inference time comparison} of different avatar reconstruction methods. All times measured on NVIDIA A100.}

\renewcommand\tabcolsep{28pt}
\resizebox{1\linewidth}{!}{
\begin{tabular}{l cc}
\toprule
\multirow{2}{*}{Method} & \multicolumn{2}{c}{Inference Time (ms)} \\
\cmidrule{2-3}
& $1024^2$ & $512^2$ \\
\midrule
% $^{\dagger}$PiCA (mesh)~\cite{ma2021pixel} & - & - \\
% $^{\dagger}$LUCAS (mesh) & - & - \\
% $^{\dagger}$LUCAS & 11.98 & 6.55 \\
% \midrule
% $^*$uPiCA (mesh) & - & - \\
% $^*$LUCAS (mesh) & - & - \\
URAvatar~\cite{li2024uravatar} & 12.84 & 6.40 \\
LUCAS & 12.01 & 6.42 \\
\bottomrule
\end{tabular}}
\label{tab:inference_time}
\end{table}

\begin{figure}[t]
\centering
\includegraphics[width=1\linewidth]{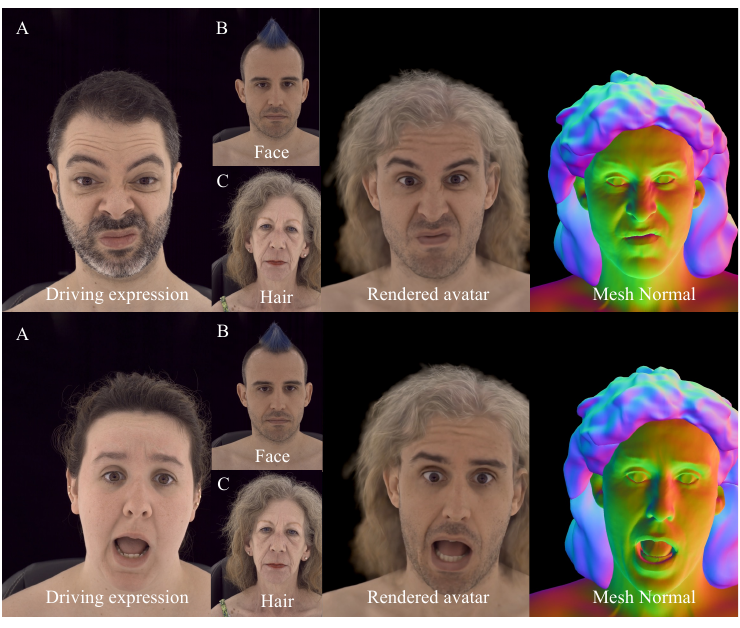}
\caption{\textbf{Application on hairstyle switching.} We combine face condition from subject B, hair condition from subject C, and expressions from subject A. Note how our model maintains high-fidelity facial details while accurately preserving the characteristics of both the chosen face and hairstyle.}
\label{fig:supp_hair_switch}
\end{figure}

\noindent \textbf{Impact of training data.} 
To understand how the number of training identities impacts our model's performance, we conducted experiments with varying numbers of subjects in the training set. Specifically, we tested our model using seven different training scales: 4, 8, 16, 32, 48, 64, and 76 identities. The model's performance shows consistent enhancement as we increase the training set size. This trend highlights the importance of diverse training samples in building robust prior representations.

\begin{figure*}[t]
\centering
\includegraphics[width=1\linewidth]{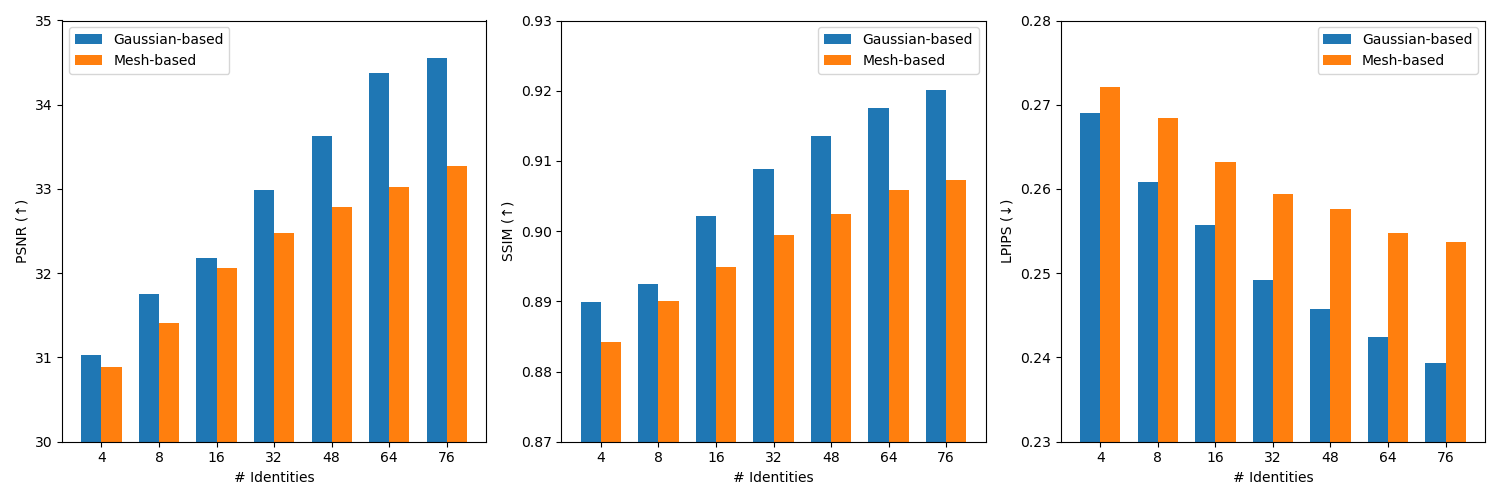}
\caption{\textbf{Ablation study} on the number of training subjects. The model's performance improves consistently with larger training sets, demonstrating the importance of diverse subjects for learning robust priors. }
\label{fig:num_ids}
\end{figure*}
\noindent \textbf{Rendering Performance.}
For all identities, we use a layered structure with separate Gaussian representations for face and hair. We experiment with two configurations: a high-quality setting using $M$ = 1024 × 1024 = 1 Mi Gaussians total (0.5 Mi each for face and hair), and a faster setting with $M$ = 512 × 512 = 0.25 Mi Gaussians total. We observe that increasing the number of Gaussians leads to quality improvement at the cost of slower rendering. As shown in Tab.~\ref{tab:inference_time}, our complete model with 1 Mi Gaussians takes 12.01 ms for rendering, while reducing to 0.25 Mi Gaussians achieves faster rendering at 6.42 ms on NVIDIA A100. Our method achieves comparable rendering speed to other Gaussian Splatting-based approaches. 
% All Gaussian-based models require 600 K iterations for convergence, which is approximately twice the training time compared to mesh-based approaches.

\section{Application}
We can independently control the hair and face appearance by using condition data from different identities. In Fig.~\ref{fig:supp_hair_switch}, we demonstrate this capability by combining the face from one subject, hair from another, and driving expressions using a third subject. Our model successfully preserves facial details like wrinkles while maintaining the distinct characteristics of the chosen face and hair styles, demonstrating its ability to decompose and recombine these elements effectively.

{
    \small
    \bibliographystyle{ieeenat_fullname}
    \bibliography{main}
}
